\documentclass[journal]{IEEEtran}
\IEEEoverridecommandlockouts
\usepackage[utf8]{inputenc}
\usepackage{cite}
\usepackage{hyperref}
\usepackage{siunitx}
\usepackage{amsmath, amssymb, amsfonts}
\usepackage{bm}
\usepackage{empheq}
\usepackage{algorithm}
\usepackage{algpseudocode}
\usepackage{graphicx}
\usepackage{subfig}
\usepackage{rotating}
\usepackage{booktabs}
\usepackage{tabularx}
\usepackage{xcolor}
\usepackage{textcomp}
\usepackage{soul} 
\usepackage{xcolor}
\usepackage{titlesec}
\usepackage{newunicodechar}
\newunicodechar{σ}{\ensuremath{\sigma}}
\usepackage{comment}
\usepackage{subfig}
\usepackage{subcaption}
\usepackage[font=small,labelfont=bf,compatibility=false]{caption}
\usepackage{xcolor}

\def\BibTeX{{\rm B\kern-.05em{\sc i\kern-.025em b}\kern-.08em%
    T\kern-.1667em\lower.7ex\hbox{E}\kern-.125emX}}

\begin{document}

\title{Secure UAV-assisted Federated Learning: A  Digital Twin-Driven Approach with Zero-Knowledge Proofs}

\author{Md Bokhtiar Al Zami,
        Md Raihan Uddin,
        Dinh C. Nguyen
\thanks{

Md Bokhtiar Al Zami, Md Raihan Uddin, and Dinh C. Nguyen are with ECE Department,  University of Alabama in Huntsville, Huntsville, AL 35899, USA, emails: (mz0024@uah.edu, mu0016@uah.edu, dinh.nguyen@uah.edu).}
}

\markboth{}%
{}

\maketitle
\begin{abstract}
Federated learning (FL) has gained popularity as a privacy-preserving method of training machine learning models on decentralized networks. However to ensure reliable operation of UAV-assisted FL systems, issues like as excessive energy consumption, communication inefficiencies, and security vulnerabilities must be solved. This paper proposes an innovative framework that integrates Digital Twin (DT) technology and Zero-Knowledge Federated Learning (zkFed) to tackle these challenges. UAVs act as mobile base stations, allowing scattered devices to train FL models locally and upload model updates for aggregation. By incorporating DT technology, our approach enables real-time system monitoring and predictive maintenance, improving UAV network efficiency. Additionally, Zero-Knowledge Proofs (ZKPs) strengthen security by allowing model verification without exposing sensitive data. To optimize energy efficiency and resource management, we introduce a dynamic allocation strategy that adjusts UAV flight paths, transmission power, and processing rates based on network conditions. Using block coordinate descent and convex optimization techniques, our method significantly reduces system energy consumption by \textcolor{black}{up to 29.6\% compared to conventional FL approaches.}
Simulation results demonstrate improved learning performance, security, and scalability, positioning this framework as a promising solution for next-generation UAV-based intelligent networks.

\end{abstract}

\section{Introduction}
Federated learning (FL) is transforming how machine learning models are trained in distributed networks. Instead of collecting and processing data at a central server, FL allows devices to train models locally and share only the learned parameters. This decentralized approach helps protect user privacy, reduce communication overhead, and improve scalability \cite{zhang2025latency,  chi2018secrecy}. As more connected devices generate vast amounts of data at the edge, FL provides an efficient way to take advantage of these data while addressing concerns about security and network congestion. In recent years, UAV (Unmanned Aerial Veichle)-assisted networks have emerged as a promising solution to support FL in dynamic and resource-constrained environments \cite{ lu2020UAV}. With their ability to move freely and establish line-of-sight (LoS) communication, UAVs can provide reliable wireless coverage in areas where traditional infrastructure is limited or unavailable\cite{banafaa2024comprehensive}. By serving as mobile edge computing (MEC) nodes, UAVs can facilitate FL by aggregating model updates from distributed devices and coordinating the training process. Despite its advantages, deploying FL in UAV networks comes with significant challenges. Limited energy supply, computational power, and storage constraints make it difficult for UAVs to sustain long-term FL operations. Additionally, the mobility of both UAVs and ground devices introduces unpredictable network conditions, leading to delays and unstable connections. Efficient scheduling and resource management are vital for reducing FL execution time and energy consumption, and for ensuring successful deployment in large-scale networks \cite{yang2019energy, yang2023joint}. Researchers are exploring new technologies such as Digital Twin (DT) to address these more efficient which has gained significant attention for its ability to create virtual replicas of physical objects or systems, offering substantial benefits in various fields such as real-time management and industrial automation \cite{sun2020reducing, zhou2024cooperative}. By simulating and optimizing the performance of physical systems in a virtual environment, DT technology provides valuable insights that can lead to improved operational efficiencies and predictive maintenance\cite{10490257, 10679522}. Recent research has focused on minimizing offloading latency in DT-enhanced edge networks, with approaches such as deep reinforcement learning (DRL) being employed to make optimal offloading decisions \cite{lu2021adaptive, mao2024dl}. This integration of DT with edge networks opens up new opportunities for more efficient and effective system management \cite{van2022digital, mao2024dl}.

Another major concern is security, even though FL avoids sharing raw data, model updates can still be vulnerable to inference attacks or unauthorized access \cite{banafaa2024comprehensive}. Addressing these issues is essential for making UAV-assisted FL practical and reliable. Recently ZKP have gained popularity as a promising approach to enhancing the security and privacy enabling verifiable model updates without exposing sensitive data \cite{zhou2024leveraging}. It's a cryptographic protocol that allows one party to prove to another that a given statement is true, without conveying any additional information apart from the fact that the statement is indeed true \cite{ballesteros2024enhancing}. Applying ZKP to FL, often referred to as zkFed, significantly boosts the security-preserving capabilities of the system \cite{wang2024zkfl, marzo2022privacy}. It ensures that during the FL process, even when data or model parameters are shared among various participants, no sensitive information is compromised \cite{10535217, shahrouz2024anonymous}. \textcolor{black}{Moreover, Qiao et al.~\cite{qiao2024transitioning} provide a comprehensive survey on transitioning from classical FL to quantum federated learning (QFL) in the IoT, highlighting the potential of quantum-enhanced privacy and computational advantages in distributed learning environments. Their work outlines how QFL can offer information-theoretic security guarantees that surpass traditional cryptographic methods, including ZKPs, especially in scenarios involving quantum communication channels and quantum data processing. Similarly, Liu et al.~\cite{liu2022blockchain} propose a blockchain-empowered FL framework tailored for healthcare-based cyber-physical systems. Their architecture integrates blockchain to ensure transparency, immutability, and decentralized trust among participating nodes, while FL preserves data privacy. This dual-layered security model addresses both data integrity and adversarial robustness, offering a complementary approach to zkFed's cryptographic proof-based validation. A comparative discussion of zkFed with such blockchain-integrated and quantum-enhanced FL frameworks would provide a more comprehensive evaluation of its security, scalability, and deployment feasibility in real-world applications.}
By enabling secure, transparent validation of data integrity and authenticity without exposing the underlying data, zkFed addresses critical security concerns in UAV-assisted networks, making it an ideal approach for scenarios where data confidentiality and security are paramount. 

\subsection{Motivations and Key Contributions}  

Despite extensive research efforts, several limitations still exist in current UAV-assisted federated FL and DT frameworks, which are highlighted below:  

\begin{itemize}  
    \item Recent works on FL in UAV networks \cite{zhang2025latency, yang2021privacy, jing2021joint} overlook the role of Digital Twin (DT) in enhancing FL efficiency, predictive maintenance, and real-time adaptation. Similarly, DT-focused studies \cite{li2023adaptive, shen2021deep} do not integrate FL, leaving untapped potential.

    \item Energy efficiency remains a key challenge. While \cite{zhang2024energy} optimizes power in UAV-based MEC networks, it lacks DT-based analytics and does not address energy trade-offs in UAV-assisted FL.

    \item Security in UAV-assisted FL is critical. Prior work \cite{yang2021privacy, lu2020UAV, qi2019UAV} focuses on encryption and differential privacy, which may reduce accuracy. Though zkFed \cite{marzo2022privacy, bamberger2022verification, lavin2024survey} offers privacy via ZKP, it remains unexplored in UAV-FL settings.

    \item UAV-FL resource optimization is limited. Existing methods optimize FL training but not joint allocation across power, compute, and mobility. DT-DRL approaches \cite{sun2020reducing, yang2023joint} improve edge computing but lack focus on FL in UAVs.
\end{itemize}

Motivated by these limitations, we propose a novel UAV-assisted zkFed framework with the following key novelties:
\begin{itemize}
    \item \textbf{Dynamic UAV Management:} We optimize UAV flight paths and resource allocation for efficient communication with ground devices.
    
    \item \textbf{Energy-Efficient Transmit Power Adjustment:} We introduce adaptive power adjustment methods to minimize energy consumption while ensuring reliable data transmission.
    
    \item \textbf{User Device Processing Rate Optimization:} We design algorithms to optimize processing rates of user devices, reducing latency and enhancing FL performance.
    
    \item \textbf{Integration of DT for Maintenance and Efficiency:} We leverage DT for real-time monitoring and predictive maintenance, enhancing system reliability.
    
    \item \textbf{Integration of zkFed for Security and Privacy:} We combine ZKP with FL to enhance security and privacy in UAV-assisted communications.
\end{itemize}



The remainder of this paper is organized as follows: Section II presents the system model, Section III describes the simulation and performance evaluation, covering the parameter settings, implementation details. Finally, Section IV concludes the paper The key acronyms and notations   are summarized in Table~\ref{table:key_acronyms} and Table~\ref{table:key_notations}, respectively.

\begin{table}[ht]
\centering
\small 
\renewcommand{\arraystretch}{1.0} 
\setlength{\tabcolsep}{3pt} 
\caption{List of Key Acronyms}
\label{table:key_acronyms}
\resizebox{\columnwidth}{!}{ 
\begin{tabular}{|p{1.2cm}|p{3.4cm}|p{1.2cm}|p{3.4cm}|}
\hline
\textbf{Acronym} & \textbf{Definition} & \textbf{Acronym} & \textbf{Definition} \\ \hline
FL     & Federated Learning & zkFed  & Zero-Knowledge Federated Learning \\ \hline
DT     & Digital Twin & UAV    & Unmanned Aerial Vehicle \\ \hline
ES     & Edge Server & DRL    & Deep Reinforcement Learning \\ \hline
ML     & Machine Learning & MD     & Mobile Device \\ \hline
IID    & Independent and Identically Distributed & FDMA   & Frequency-Domain Multiple Access \\ \hline
AWGN   & Additive White Gaussian Noise & SNARK  & Succinct Non-interactive Argument of Knowledge \\ \hline
ZKP    & Zero-Knowledge Proof & CPU    & Central Processing Unit \\ \hline
SQP    & Sequential Quadratic Programming & MEC & Mobile Edge Computing \\ \hline
\end{tabular}}
\end{table}

\begin{table}[ht]
\centering
\small 
\renewcommand{\arraystretch}{1.0} 
\setlength{\tabcolsep}{3pt} 
\caption{List of Key Notations}
\label{table:key_notations}
\resizebox{\columnwidth}{!}{ 
\begin{tabular}{|p{1.5cm}|p{3.5cm}|p{1.5cm}|p{3.5cm}|}
\hline
\textbf{Notation} & \textbf{Definition} & \textbf{Notation} & \textbf{Definition} \\ \hline
$N$   & Number of user devices & $K$   & Number of global rounds \\ \hline
$D_n$ & Local dataset of user $n$ & $f_n$ & Computation frequency of user $n$ \\ \hline
$x[k], y[k]$ & UAV horizontal position & $H$ & UAV flight altitude \\ \hline
$L$   & Max UAV flight per slot & $v_{\text{max}}$ & Max UAV velocity \\ \hline
$T_{\text{max}}$ & Max allowable latency & $p_n$ & Transmission power of user $n$ \\ \hline
$p_{\text{UAV}}$ & UAV transmission power & $\sigma^2$ & AWGN power \\ \hline
$g_{\text{UAV}}[k]$ & Channel gain at slot $k$ & $d_{n,UAV}[k]$ & Distance between user $n$ and UAV \\ \hline
$B$   & Bandwidth per user & $R_n^{\text{up}}[k]$ & Uplink data rate at slot $k$ \\ \hline
$R_n^{\text{down}}[k]$ & Downlink data rate & $I_n$ & Local iterations at user $n$ \\ \hline
$C_n$ & CPU cycles per sample & $E_n^{\text{train}}$ & Training energy of user $n$ \\ \hline
$E_{\text{com}}^{\text{up}}$ & Uplink energy consumption & $E_{\text{com}}^{\text{down}}$ & Downlink energy consumption \\ \hline
$T_n^{\text{train}}$ & Training time of user $n$ & $T_{\text{com}}^{\text{up}}$ & Uplink latency \\ \hline
$T_{\text{com}}^{\text{down}}$ & Downlink latency & $T_n^{\text{total}}$ & Total latency per round \\ \hline
$\alpha$ & Capacitance coefficient & $\kappa$ & Hardware efficiency coefficient \\ \hline
$\beta_0$ & SNR scaling factor & $\phi_n$ & Processing rate of user $n$ \\ \hline
$\lambda$ & Slack variable & $\xi, \eta, \Lambda$ & Slack variables for UAV optimization \\ \hline
\end{tabular}}
\end{table}

\section{System Model}

\subsection{Overall System Architecture}
\begin{figure}[ht]
    \centering
    \includegraphics[width=0.48\textwidth]{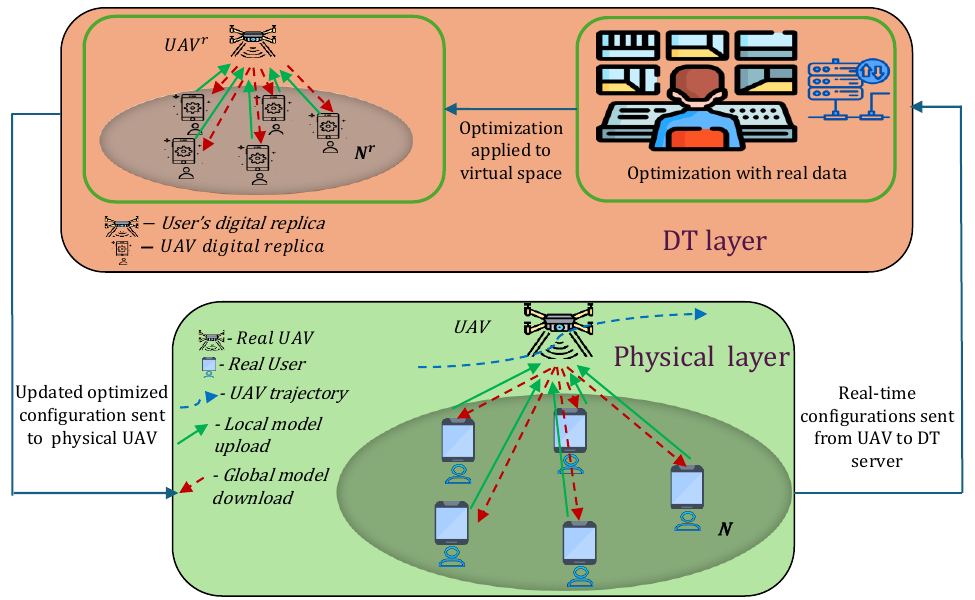}
    \caption{The proposed zkFed-enabled UAV-assisted FL system architecture.}
    \label{fig:system_architecture}
\end{figure} 

Our proposed DT-assisted zkFed framework for UAV-enabled privacy-preserving FL is illustrated in Fig.~\ref{fig:system_architecture}. The system integrates UAVs, FL agents (user devices), and a DT layer to enhance energy efficiency, optimize resource allocation, and ensure robust privacy and security. The UAVs function as aerial FL aggregators, dynamically adjusting their flight trajectories, transmission power, and user association strategies to maintain efficient communication and resource allocation. We consider a UAV assisted FL system with \(N\) user devices, denoted as \(\mathcal{N} = \{1, 2, ..., N\}\), each of which possesses a private dataset \(D_n\). The system operates over \(K\) global communication rounds, with each device training its local model using dataset \(D_n\) and uploading encrypted model parameters to a UAV based aggregator. Given the inherent constraints of UAV assisted networks, including limited energy capacity, dynamic mobility, and constrained communication resources, we employ a DT model to enable real-time optimization of computational efficiency and network performance.

The overall interactions between UAVs, the DT framework, and zkFed-based FL can be summarized as follows:

\begin{itemize}
    \item User devices: The user devices (edge clients) are responsible for performing local FL model training on their private datasets \(D_n\). To ensure energy-efficient training, they utilize computation resources, specifically CPU cycles, to optimize processing rates. Once the training is completed, the encrypted model updates are transmitted to UAVs for aggregation using frequency-domain multiple access (FDMA). Additionally, these user devices serve as digital replicas in the DT layer, enabling real-time performance evaluation and optimization.

    \item UAV aggregators: UAVs act as mobile FL coordinators, dynamically adjusting their trajectory, transmit power, and computation allocation to optimize communication and learning efficiency. They utilize adaptive energy-aware scheduling to balance FL model training with UAV operational constraints, ensuring optimal resource utilization. To enhance security and privacy, UAVs implement zkFed, enabling the verification of encrypted model updates without exposing sensitive data. Furthermore, they maintain bidirectional communication with the DT layer, receiving optimized flight trajectories and system control recommendations to improve overall system performance.
    
    \item DT layer: The DT layer constructs virtual models of UAVs and user devices, enabling the simulation of real-time system performance. It provides predictive optimization for UAV trajectory planning, resource allocation, and energy consumption, ensuring efficient operations. Additionally, the DT layer continuously analyzes communication latencies, processing delays, and security vulnerabilities, refining FL operations to enhance overall system reliability and security.
    
     \item Zero-Knowledge Proof for Federated Learning (zkFed): 
     By having clients train local models using local data and then submit those models to the server rather than the raw data, FL systems seek to protect client privacy. The server then combines these local models to provide an updated global model. FL systems typically involve a central server that, while generally trustworthy, operates under an \textcolor{black}{``honest-but-curious''} model where it can access all client data. To improve privacy, these systems can incorporate techniques like adding noise to the data \cite{nguyen2021federated}. Despite the advanced features of FL frameworks, clients must trust that the server accurately aggregates the data. However, this aggregation is often opaque, posing risks such as the potential for a global aggregation attack. In such scenarios, the server might not utilize all client data to reduce computational demands or unfairly prioritize some client models over others \cite{sharma2024review}.

\end{itemize}

\noindent\textcolor{black}{In our framework, the DT layer maintains synchronization with the physical UAV--user network through continuous telemetry and computation-status updates. Each UAV is equipped with a GPS receiver and inertial sensors for real-time position and motion tracking, along with wireless link-quality estimators to monitor channel gain as expressed in \ref{eq:3}. Ground user devices periodically report their current computation frequency, residual energy, and training status, which are represented in the physical model by parameters such as \(f_n\), \(E^{\mathrm{train}}_n\), and \(T^{\mathrm{train}}_n\). These measurements are transmitted to the DT server via the existing UAV--ground FDMA links and relayed over the UAV--cloud backhaul using lightweight IoT messaging protocols to minimize overhead. UAV telemetry and channel-state information are refreshed every 1--2 seconds, while aggregated user computation and energy metrics are updated every 5 seconds. Upon receiving these updates, the DT layer refreshes the virtual replica of the network and executes optimization routines to adjust UAV trajectory \((x[k], y[k])\), transmission powers \(q_n\) and \(q_{\mathrm{UAV}}[k]\), and user scheduling. The optimized control parameters are then returned to the UAV with an end-to-end feedback delay of less than 200~ms, allowing DT-driven adjustments to be applied within the same global FL round.} Each FL round consists of four key stages, as illustrated in Fig.~\ref{fig:system_architecture}:

\subsubsection{Local model training and DT-optimized computation scheduling}   
    Each user performs local FL model training using dataset \(D_n\) and optimizes its processing rate (\(f_n\)) to minimize energy consumption. The DT layer simulates user computational loads, predicting energy consumption and scheduling adjustments. To reduce communication latency, UAVs dynamically allocate bandwidth (\(B\)) and transmit power (\(p_n\)) to ensure reliable model updates.

\subsubsection{ UAV-assisted model aggregation and zkFed verification}  
    After local training, user devices encrypt their model parameters using homomorphic encryption and transmit them to the UAV using frequency-domain multiple access (FDMA). The UAV aggregates these encrypted models using a privacy-preserving zkFed scheme, verifying each update via zero-knowledge proofs (ZKPs). The DT layer continuously monitors UAV trajectory adjustments, ensuring optimal positioning for minimal transmission delays.

\subsubsection{ Global model update and secure zk-SNARK validation}  
    The UAV updates the global FL model by aggregating verified encrypted updates. A zk-SNARK proof is generated, allowing devices to verify the validity of the global model without revealing the underlying computations. The DT simultaneously recomputes UAV trajectory optimizations, dynamically adjusting flight paths based on real-time user density and communication constraints.

\subsubsection{Model distribution and next iteration}  
    The UAV broadcasts the updated global model to all user devices. Users validate the model using zkFed-based authentication mechanisms, ensuring the integrity of the training process. The DT continuously optimizes energy allocation, trajectory adjustments, and user scheduling to improve efficiency in the subsequent learning round.\

\subsection{Physical System Model}
We consider a DT-assisted FL framework where a single UAV works as an aerial aggregator to collaborate with a group of terrestrial users (denoted as by \( \mathcal{N} = \{1, 2, \ldots, N\} \)) within its operational range. It is assumed the UAV's position is denoted as (\( x[k], y[k], H\)). the user $n$'s position is denoted  as \( [x^E_{\textit{n}}, y^E_{\textit{n}}, 0] \). The UAV is assumed to fly at a fixed altitude \( H \) above ground, which is considered the minimum height necessary to avoid obstacles.

The UAV's operation time can be partitioned into small time slots in the operational time span \( T \) into \( K \) equal intervals, resulting in \( T = K \Delta t \), where \( \Delta t \) implies the duration of each interval. This segmentation granularity allows us to treat the UAV's position as stable within each interval, denoted by (\( x[k], Y[k], H \)). The UAV's horizontal trajectory (\( x(t), y(t) \)) across \( T \) can be succinctly represented by the sequence \( \{x[k], y[k]\}_{k=1}^{K} \). Given the UAV's maximum velocity \( v_{\textit{max}} > 0 \), its traversal distance per interval stands at \( L = v_{\textit{max}} \Delta t \).

The UAV's initial and final positions, defined as (\( x_0, y_0, H \)) and (\( x_F, y_F, H \)) respectively, align with mission directives. As such, the UAV's navigational constraints are articulated as follows:

\begin{subequations}
\begin{align}
    &(x[1] - x_0)^2 + (y[1] - y_0)^2 \leq L^2, \\
    &(x[k + 1] - x[k])^2 + (y[k + 1] - y[k])^2 \leq L^2, \nonumber \\
    &\qquad\qquad k = 1, \ldots, K - 1  \\
    &(x_F - x[K])^2 + (y_F - y[K])^2 \leq L^2.
\end{align}
\end{subequations}

Within this framework, the UAV assumes the role of a parameter server, facilitating each user \( n \) in training a FL model using their unique local dataset \( D_{\textit{n}} \), which collectively enriches the overarching dataset \( D \). Comprising input-output pairs \( \{(\mathbf{v}_{\textit{n}}, z_{\textit{n}})\}_{i=1}^{D_{\textit{n}}} \), each local dataset \( D_{\textit{n}} \) underscores the collaborative pursuit of model enhancement while steadfastly safeguarding data privacy concerns.


The system starts with local model training, in this process the computation frequency \( \phi_{\textit{n}} \) of user \( n \) determines the time \( T_{\textit{n}}^\textit{train} \) required for processing data  and the energy consumption \( E_{\textit{n}}^{train} \) for completing \( C_{\textit{n}}D_{\textit{n}} \) CPU cycles at user \( n \) is given by
\begin{equation}
     E_{\textit{n}}^{train} = \alpha I_{\textit{n}}C_{\textit{n}}D_{\textit{n}}\phi_{\textit{n}}^2,
\end{equation}
where \( \alpha \) is the effective capacitance coefficient.

After local computations, each user uploads their local FL model using frequency domain multiple access (FDMA). In this step, the UAV communicates with terrestrial users primarily through Line-of-Sight (LoS) links. The power gain of the channel from the UAV to a user during the \( k \)-th time slot can be described by the free-space path loss model:

\begin{equation}
    g_{UAV}[k] = \frac{\alpha_0}{d_{n, UAV}^{2}[k]}  =\frac{\alpha_0}{ x[k]^2 + y[k]^2 + H^2}, \label{eq:3}
\end{equation}
where \( \alpha_0 \) represents the power gain at a reference distance \( d_0 = 1 \) meter, which is influenced by the carrier frequency and the antenna gains of both the transmitter and the receiver. Moreover, \(d_{\textit{UAV, n}}[k]\) refers to the distance from the UAV to a terrestrial user at time slot $k$, which is denoted by \(d_{\textit{n, UAV}}[k] = \sqrt{x[k]^2 + y[k]^2 + H^2}\).
The users transmit power at time slot \( k \), represented as \( p_{\textit{n}}[k] \), is subject to both average and peak power constraints, denoted by \( \bar{p} \) and \( p_{\textit{peak}} \) respectively, i.e.,
\begin{align}
\frac{1}{K} \sum_{k=1}^{K} p_{\textit{n}}[k] &\leq \bar{p}_{\textit{n}}, \\
0 &\leq \sum_{k=1}^K p_{\textit{n}}[k] \leq P_{\textit{n}}, \quad \forall k
\end{align}


\begin{subequations}
\begin{align}
    &\frac{1}{K} \sum_{k=1}^{K} p_{\textit{n}}[k] \leq \bar{p}_{\textit{n}}, \\
    &0 \leq \sum_{k=1}^K p_{\textit{n}}[k] \leq P_{\textit{n}}, \quad \forall k.
\end{align}
\end{subequations}

To ensure these constraints are meaningful, it is assumed that \( \bar{p} \) is less than \( p_{\textit{peak}} \).


Now we can calculate he uplink data rate \( R_{\textit{up}} \) for user \( n \) is determined by

\begin{equation}
    R^{\textit{up}}_{\textit{n}}[k] = B \log_2 \left(1 + \frac{\alpha_0 p_{\textit{n}}[k]}{\sigma^2 d_\textit{n,UAV}^{2}[k]}\right),
\end{equation}   
where \( B\) is bandwidth allocated to user (it is assumed bandwidth is equal to every user) and \(\sigma^2\) is the Additive White Gaussian Noise (AWGN) power. Similarly, the downlink data rate \( R_{\textit{down}} \) for broadcasting the global model is given by
\begin{equation}
    R^{\textit{down}}_{\textit{n}}[k]  = B \log_2 \left(1 + \frac{\alpha_0 p_{\textit{UAV}}[k]}{\sigma^2 d_\textit{UAV, n}^2[k]}\right),
\end{equation}
where \(p_{\textit{UAV}}[k]\) is the UAV's transmit power at time slot $k$.

\subsection{DT Model}
DT technology creates a detailed replica of physical systems, incorporating hardware setups, historical data, and real-time operational statuses. This functionality allows for seamless interaction with the physical system through mechanisms that enable real-time updates and control. The DT model for UAV-assisted FL can be defined as 
\begin{equation}
    DT = (\mathcal{N}, \mathcal{N}^{r}), (UAV, UAV^{r}) .
\end{equation}

Digital replicas of the physical network, referred to as \( N^{r} \) and \( UAV^{r} \), include all users and the UAV. By utilizing real-time updated data from physical entities, the digital services within DT layer provide a wide range of functionalities that enable autonomous management of the system.



The DT model \( DT_{\textit{n}} \), employed for the local training process of the \( n \)-th user, can be expressed as $DT_{\textit{n}} =  (\Tilde{\phi}_{\textit{n}} , \phi'_{n}),$
where \( \Tilde{\phi}_{\textit{n}} \) denotes the estimated processing rate of the physical user, and \( \phi'_{n} \) represents the deviation between the physical user and its DT model representation. The estimated time required to execute the task locally is given by
\begin{equation}
    \Tilde{T}_{\textit{n}}^{\textit{train}} = \frac{C_{\textit{n}}I_{\textit{n}}D_{\textit{n}} }{\Tilde{\phi}_{\textit{n}}}.
\end{equation}

Assuming that the deviation between the physical user and its digital representation can be identified in advance, the latency gap for local model training can be calculated as follows:
\begin{equation}
    \Delta T_{\textit{n}}^{\textit{train}} = \frac{C_{\textit{n}}I_{\textit{n}}D_{\textit{n}}\phi'_{n}}{\Tilde{\phi}_{\textit{n}}(\Tilde{\phi}_{\textit{n}} - \phi'_{n})}.
\end{equation}
Hence we can write the actual time and energy uses for local model training as
\begin{equation}
    T_{\textit{n}}^{\textit{train}} = \Tilde{T}_{\textit{Loc}}^{n} + \Delta T_{\textit{n}}^{\textit{train}} = \frac{C_{\textit{n}}I_{\textit{n}}D_{\textit{n}} }{\Tilde{\phi}_{\textit{n}}- \phi'_{n}}.
\end{equation}

\begin{equation}
    E_{\textit{n}}^{train} = \alpha I_{\textit{n}}C_{\textit{n}}D_{\textit{n}}(\Tilde{\phi}_{n} - \phi'_{n})^2.
\end{equation}

\subsubsection{Local Model Uploading}

The uploading power \( \textit{DT}_{\textit{Pn}}\) required for the local model upload from the \( n \)-th user to the UAV can be expressed as $\textit{DT}_{\textit{Pn}} = (\Tilde{p}_{\textit{n}}, p'_{\textit{n}}),$
where \( \Tilde{p}_{\textit{n}}[k] \) is the estimated power and \( p'_{\textit{n}}[k] \) represents the deviation between the physical system and its DTmodel. The estimated time to upload the local model, based on the DTmode estimation, is given by

\begin{equation}
     \Tilde{T}_{\textit{n}}^{\textit{up}}[k] = \frac{D_{\textit{n}}}{R_{up}} =\frac{D_{\textit{n}}}{B \log_2 \left(1 + \frac{\beta \Tilde{p}_{\textit{n}}}{ d_\textit{n,UAV}^{2}[k]}\right)},
\end{equation}
where $\beta_{0} = \frac{\alpha_0}{\sigma^2}$. To incorporate the deviation \( p'_{\textit{n}}\), the actual transmission power becomes \( \Tilde{p}_{\textit{n}} - p'_{\textit{n}} \). Therefore, the actual time to upload the local model is given by:

\begin{equation}
     T_{\textit{com}}^{\textit{up}}[k] = \frac{D_{\textit{n}}}{R_{up}} =\frac{D_{\textit{n}}}{B \log_2 \left(1 + \frac{\beta_{0} (\Tilde{p}_{\textit{n}} - p'_{\textit{n}})}{d_\textit{n,UAV}^{2}[k]}\right) } .
\end{equation}





Thus we can write the actual energy consumption for local model uploading as

\begin{equation}
    E_{\textit{com}}^{\textit{up}}[k] = (\tilde{p}_{\textit{n}}-p'_{\textit{n}})\cdot T_{\textit{com}}^{\textit{up}}[k].
\end{equation}









\subsubsection{Global model downloading}

Transmit power associated with the UAV, denoted as \( \textit{DT}_{p_{\textit{UAV}}} \), plays a crucial role in the downloading process within the DT model representation, denoted by $\textit{DT}_{p_{\textit{UAV}}} = (\Tilde{p}_{\textit{UAV}}[k], p'_{\textit{UAV}}[k]),$
where \( \Tilde{p}_{\textit{UAV}} \) denotes the estimated transmit power of UAV, while \( p'_{\textit{UAV}} \) signifies the divergence between the real UAV and its DT model counterpart. The estimated time for fetching the global model can be computed as

\begin{equation}
     \Tilde{T}_{\textit{n}}^{\textit{down}}[k] =\frac{w_{\textit{n}}}{B \log_2 \left(1 + \frac{\alpha_0 \Tilde{p}_{\textit{UAV}}[k]}{\sigma^2 d_\textit{UAV, n}^{2}[k]}\right) } .
\end{equation}





To compensate for deviations in transmission power \( p'_{\textit{UAV}}[k] \), the operational transmission power \( \Tilde{p}_{\textit{UAV}}[k] \) is adjusted accordingly:

\begin{equation}
     T_{\textit{com}}^{\textit{down}}[k] = \frac{w_{\textit{n}}}{B \log_2 \left(1 + \frac{\alpha_0 (\Tilde{p}_{\textit{UAV}}[k] - p'_{\textit{UAV}}[k])}{\sigma^2 d_{\textit{UAV}, n}^{2}[k]}\right) },
\end{equation}

where \( T_{\textit{com}}^{\textit{down}}[k] \) represents the actual time required for retrieving the global model considering adjusted transmission power.



The actual energy consumption associated with retrieving the global model can be inferred as follows:



\begin{equation}
    E_{\textit{com}}^{\textit{down}}[k] = (\Tilde{p}_{\textit{UAV}}[k] - p'_{\textit{UAV}}[k]) \cdot T_{\textit{com}}^{\textit{down}}[k].
\end{equation}




\subsubsection{Total Latency and Energy Consumption}

Now we can calculate the total latency in DT for each global round is given by:

\begin{equation}
    T_{n}^{total} = \max_{\forall n \in N}(T_{\textit{n}}^{\textit{train}} + T_{\textit{com}}^{\textit{up}}[k] + T_{\textit{com}}^{\textit{down}}[k]).
\end{equation}

This equation computes the maximum total latency across all users in the system. It considers the local model training time \(T_{\textit{n}}^{\textit{train}} \), uplink communication time \( T_{\textit{com}}^{up} \), and downlink communication time \( T_{\textit{com}}^{down} \).

And the energy consumption is given by

\begin{equation}
    E_{n}^{total} = \sum_{n=1}^N (E_{\textit{n}}^{train} + E_{\textit{com}}^{\textit{up}} + E_{\textit{com}}^{\textit{down}}[k]).
\end{equation}

This equation computes the total energy consumption for each user \( n \) in the system. It sums up the energy consumed during local model training \( E_{\textit{n}}^{train} \), uplink communication \( E_{\textit{com}}^{up} \), and downlink communication \( E_{\textit{com}}^{down} \).


\subsection{Problem Formulation}
Thus, the goal of the DT-assisted FL system is to minimize the energy consumption of the physical FL systems with the assistance of the DT. The problem can be formulated as follows\cite{chi2018secrecy}:

\begin{subequations}
\begin{align}
    \begin{split}
    \min_{\substack{x[k], y[k], \\ \Tilde{\Phi_{\textit{n}}}, \Tilde{p_{\textit{n}}}, \tilde{p_\textit{UAV}[k]}}} \quad & \sum_{n=1}^N \Bigg( \alpha I_{\textit{n}} C_{\textit{n}} D_{\textit{n}} (\Tilde{\phi}_{n} - \phi'_{n})^2 \\
    & \quad + \frac{(\tilde{p}_{\textit{n}} - p'_{\textit{n}}) \cdot D_{\textit{n}}}{B \log_2 \left(1 + \frac{\beta (\Tilde{p}_{\textit{n}} - p'_{\textit{n}})}{d_\textit{n,UAV}^{2}[k]}\right) } \\
    & \quad + \frac{(\Tilde{p}_{\textit{UAV}}[k] - p'_{\textit{UAV}}[k]) \cdot w_{\textit{n}}}{B \log_2 \left(1 + \frac{\alpha_0 (\Tilde{p}_\textit{UAV}[k] - p'_\textit{UAV}[k])}{\sigma^2 d_\textit{UAV, n}^{2}[k]}\right) } \Bigg)
    \end{split}  \label{eqn:22} \\
    \textit{s.t.} \quad & 0 \leq \phi_{n} \leq \phi_{\textit{max}}, \quad \forall n \label{eq:sub22a}\\
    & 0 \leq p_{\textit{n}} \leq p_{\textit{max}}, \quad \forall n \label{eq:sub22b}\\
    & 0 \leq p_{UAV} \leq p_{\textit{UAV, max}}[k] \label{eq:sub22c}\\
    & T_{n}^{total}\leq T_{\textit{max}}, \quad \forall n \label{eq:sub22d}\\
    & (x_F - x[K])^2 + (y_F - y[K])^2 \leq L^2 \label{eq:sub22e}
\end{align}
\end{subequations}

This problem incorporates constraints on bandwidth, computation frequency, upload power, download power, and latency for our system model. Constraint \eqref{eq:sub22a} defines the permissible range for processing rate \( \phi_{\textit{n}} \) for each user. Constraint \eqref{eq:sub22b} limits the transmission power \( p_{\textit{n}} \) for each user within the maximum power \( p_{\textit{max}} \). Constraint \eqref{eq:sub22c} ensures the UAV's transmission power \( p_{UAV} \) does not exceed the maximum power \( p_{\textit{UAV, max}} \). Constraint \eqref{eq:sub22d} guarantees that the total time \( T_{\textit{n}} \), including local computation and communication times, does not surpass the maximum allowable time \( T_{\textit{max}} \). Constraint \eqref{eq:sub22e} ensures that the final position of the UAV is within a permissible region defined by radius \( L \) centered at \((x_F, y_F)\).

 For our Problem, we assume simplified variables to manage complexity. First we assume the actual computation frequency of user in DT defined as \( f_{\textit{n}}= \Tilde{\phi}_{n} - \phi'_{n}\). Similarly actual transmission power for user \( n \), \( q_{n}[k] = \tilde{p}_n - p'_n \), and actual UAV's transmission power, expressed as \( q_{UAV}[k] = \tilde{p}_{UAV}[k] - p'_{UAV}[k] \). The constants \( \beta = \frac{D_n}{B} \) and \( \gamma = \frac{w_n}{B} \) simplify bandwidth-related considerations for users and the UAV respectively. Now we can write the uapdated problem as,

\textit{\pmb{Problem 1:}}
\begin{subequations}
\begin{align}
\begin{split}
    \min_{\substack{x[k], y[k], \\ f_{n}, q_{n}, q_{UAV}[k]}} \quad & \sum_{n=1}^N \Bigg( \alpha I_n C_n D_n f_{n}^2\\
    & \quad + \frac{q_n \cdot \beta}{\log_2 \left(1 + \frac{\beta q_n}{d_{n,UAV}^{2}[k]}\right) } \\
    & \quad + \frac{q_{UAV}[k] \cdot \gamma}{ \log_2 \left(1 + \frac{\alpha_0 q_{UAV}[k]}{\sigma^2 d_{UAV, n}^{2}[k]}\right) } \Bigg)\end{split}  \label{eqn:23} \\
    \textit{s.t.} \quad & 0 \leq f_{n} \leq f_{\textit{max}}, \quad \forall n \\
    & 0 \leq q_{n} \leq q_{\textit{max}}, \quad \forall n \\
    & 0 \leq q_{UAV}[k] \leq q_{\textit{UAV, max}}[k] \\
    & T_n^{\text{total}} \leq T_{\textit{max}}, \quad \forall n \label{eqn:23d}\\
    & (x_F - x[K])^2 + (y_F - y[K])^2 \leq L^2 \label{eqn:23e} 
\end{align}
\end{subequations}

\subsection{Proposed Solution}

It is evident that the objective function \eqref{eqn:23} as well as the constraints \eqref{eqn:23d} and \eqref{eqn:23e} are non-convex. Due to the complexity of directly solving \textbf{\textit{Problem 1}}, we decompose it into two more manageable subproblems. Specifically, we divide the original problem into \textit{\textbf{Subproblem 1}}, which optimizes parameters for the users, and \textit{\textbf{Subproblem 2}}, which focuses on optimizing the UAV parameters.


\textit{\pmb{Subproblem 1:}}
\begin{subequations}
\begin{align}
\begin{split}
    \min_{\substack{f_{\textit{n}}, q_{\textit{n}}}} \quad & \sum_{n=1}^N \Bigg( \alpha I_n C_n D_n f_{n}^2 \\
    & \quad + \frac{q_n \cdot \beta}{\log_2 \left(1 + \frac{\beta_{0} q_n}{d_{n,UAV}^{2}[k]}\right) } \\
    & \quad + E_{\textit{com}}^{\textit{down}}[k] \Bigg) \end{split} \label{eqn:24} \\
    \textit{s.t.} \quad & 0 \leq f_{\textit{n}} \leq f_{\textit{max}}, \quad \forall n \label{eqn:24a}\\
    & 0 \leq q_{\textit{n}} \leq q_{\textit{max}}, \quad \forall n \label{eqn:24b}\\
    & T_{n}^{total}\leq T_{\textit{max}}, \quad \forall n \label{eq:24c} 
\end{align}
\end{subequations}


\subsubsection{Convexification of Sub-problem 1}

Subproblem 1 is non-convex due to the objective function \eqref{eqn:24} and the constraint \eqref{eq:24c}. In order to convexify \eqref{eqn:24}, we introduce a slack variable \(z_{\textit{n}}\) as
\begin{equation}
    z_{\textit{n}} \geq \frac{q_{\textit{n}} \beta}{\log_2 \left(1 + \frac{q_{\textit{n}}}{d_\textit{n,UAV}^{2}[k]}\right)}.
    \label{eq:25}
\end{equation}

\noindent Now we introduce more slack variables \( \psi \) and \( \theta \) such that:



\begin{subequations}
\begin{empheq}[left=\eqref{eq:25} \quad \boldsymbol{\Longleftrightarrow} \quad \left\{ \begin{aligned},right={\end{aligned}\right.}]{align}
  & \psi z_{\textit{n}} \geq \beta q_{\textit{n}} \label{eq:sub26a},\\
  & \log_2 (1 + \theta) \geq \psi \label{eq:sub26b} \\
  & \frac{q_{\textit{n}}}{d_{\textit{n},UAV}^{2}[k]} \geq \theta \label{eq:sub26c}
\end{empheq}
\end{subequations}

Here we can see \eqref{eq:sub26a} and \eqref{eq:sub26b} are non-convex. We can covexify these using Taylor expansion and inequalities respectively,

\begin{align}
\textbf{\underline{\eqref{eq:sub26a}:}} \quad \boldsymbol{\Rightarrow} \quad & \psi z_{\textit{n}} \geq \beta q_{\textit{n}}, \notag \\
\boldsymbol{\Rightarrow} \quad & \frac{1}{4}(z_{\textit{n}}+\psi)^2 - \frac{1}{4}(z_{\textit{n}}-\psi)^2 \geq \beta q_{\textit{n}}, \notag \\
\boldsymbol{\Rightarrow} \quad & \frac{1}{4}(z_{\textit{n}}+\psi)^2 - \beta q_{\textit{n}}  \geq \frac{1}{4}(z_{\textit{n}}-\psi)^2. \label{eq:27}
\end{align}

In \eqref{eq:27}, the RHS is convex. For LHS, we can use the first order Taylor expansion:
\begin{equation}
\begin{aligned}
    (z_{\textit{n}}+\psi)^2 \geq & \ (z_{\textit{n(i)}}+\psi_{\textit{(i)}})^2 + \\
    &  2(z_{\textit{n(i)}}+\psi_{\textit{(i)}})(z_{\textit{n}}+\psi - z_{\textit{n(i)}}-\psi_{\textit{(i)}}) \label{eq:28}
\end{aligned}
\end{equation}
So we can rewrite \eqref{eq:28} as

\begin{equation}
\begin{aligned}
    \frac{1}{4}[(z_{\textit{n(i)}}+\psi_{\textit{(i)}})^2 & + 2(z_{\textit{n(i)}}+\psi_{\textit{(i)}})(z_{\textit{n}}+\psi - z_{\textit{n(i)}}-\psi_{\textit{(i)}})] \\
    & - \beta q_{\textit{n}} \geq \frac{1}{4}(z_{\textit{n}}-\psi)^2. \label{eq:29}
\end{aligned}
\end{equation}

\textbf{\underline{\eqref{eq:sub26b}:}} Here we use the inequality to convexify:

\begin{equation*}
\begin{aligned}
    \ln(1+z) & \geq \ln(1+z_{\textit{i}}) + \frac{z_{\textit{i}}}{z_{\textit{i}+1}} - \frac{z_{\textit{i}}^2}{z_{\textit{i}+1}} \frac{1}{z},
\end{aligned}
\end{equation*}
thus the equation \eqref{eq:sub26b} is approximated as follows:
\begin{align}
 \quad &  \ln(1+\theta_{\textit{i}}) + \frac{\theta_{\textit{i}}}{\theta_{\textit{i}+1}} - \frac{\theta_{\textit{i}}^2}{\theta_{\textit{i}+1}} \frac{1}{\theta} \geq \psi\ln2. \label{eq:30}
\end{align}
\textbf{\underline{\eqref{eq:sub26c}:}} Equation \eqref{eq:sub26c} is inherently convex, allowing for direct solution using CVX.

\textbf{\underline{Convexify of constraints \eqref{eq:24c}:}}

\begin{equation}
\begin{aligned}
&\Rightarrow T_{n}^{\text{total}} \leq T_{\textit{max}}, \forall n \\
&\Rightarrow T_{n}^{\text{train}} + T_{\text{com}}^{\text{up}}[k] + T_{\text{com}}^{\text{down}}[k] \leq T_{\textit{max}}, \forall n \\
&\Rightarrow \frac{C_{n} I_{n} D_{n}}{f_{n}} + \frac{\beta}{\log_2 \left(1 + \frac{\beta_{0} q_{n}}{d_{n,UAV}^{2}[k]}\right)} + 
 T_{\text{com}}^{\text{down}}[k] \leq T_{\textit{max}}, \forall n \label{eq:31}
\end{aligned}
\end{equation}

From \eqref{eq:31} we can see that the first term of the LHS is convex, bout 2nd term is non-convex. Using slack varible $z_{\textit{n}}$ here we get  
    \begin{equation}
\begin{aligned}
\frac{C_{n} I_{n} D_{n}}{f_{n}} + \frac{z_{\textit{n}}}{q_{\textit{n}}} + 
 T_{\text{com}}^{\text{down}}[k] \leq T_{\textit{max}}, \forall n \label{eq:32}
\end{aligned}
\end{equation}
To convexify the non-convex term in equation \eqref{eq:32}, we use the slack variable \(z_n\) as introduced previously. The term \(\frac{z_n}{q_n}\) needs to be convexified:

We can rewrite the term \(\frac{z_n}{q_n}\) as:
    \begin{equation}
         \frac{z_n}{q_n} \leq \frac{1}{4} \left(\left(z_n + \frac{1}{q_n}\right)^2 - \left(z_n - \frac{1}{q_n}\right)^2\right),
    \end{equation}

where 2nd quadratic term of the RHS is convex. Applying the first-order Taylor expansion for the 1st quadratic term as:
\begin{equation}
\begin{aligned}
    \left(z_n + \frac{1}{q_n}\right)^2 &\geq (z_{n(i)} + \frac{1}{q_{n(i)}})^2 \\
    &\quad + 2(z_{n(i)} + \frac{1}{q_{n(i)}}) \\
    &\quad \quad  \left(z_n + \frac{1}{q_n} - z_{n(i)} - \frac{1}{q_{n(i)}}\right). \label{eq:33}
\end{aligned}
\end{equation}\\
Hence we can rewrite the subproblem 1 as follows:
\textit{\pmb{Reformulated Subproblem 1:}}
\begin{subequations}
\begin{align}
\begin{split}
    \min_{\substack{f_{\textit{n}}, q_{\textit{n}}}} \quad & \sum_{n=1}^N \Bigg( \alpha I_n C_n D_n f_{n}^2 + z_{\textit{n}}\ + E_{\textit{com}}^{\textit{down}}[k] \Bigg) \end{split} \label{eqn:34} \\
    \textit{s.t.}  \quad & \eqref{eq:33}, \eqref{eq:32},\eqref{eq:30},\eqref{eq:29}, \eqref{eq:sub26c}, \eqref{eqn:24a} - \eqref{eqn:24b}.\label{eq:sub34}
\end{align}
\end{subequations}\\


\textit{\pmb{Subproblem 2:}}
\begin{subequations}
\begin{align}
\begin{split}
    \min_{\substack{x[k], y[k], q_\textit{UAV}[k]}} \quad & \sum_{n=1}^N \Bigg( E_{\textit{n}}^{train} + E_{\textit{com}}^{\textit{up}}[k] \\
    & \quad + \frac{q_{UAV}[k] \cdot \gamma}{ \log_2 \left(1 + \frac{\alpha_0 q_{UAV}[k]}{\sigma^2 d_{UAV, n}^{2}[k]}\right) } \Bigg) \end{split} \label{eq:35} \\
    \textit{s.t.} \quad 
    & 0 \leq q_\textit{UAV}[k] \leq q_{\textit{UAV, max}}[k] \label{eq:sub35a} \\
    & (x_F - x[K])^2 + (y_F - y[K])^2 \leq L^2 \label{eq:sub35b}
\end{align}
\end{subequations}
\subsubsection{Convexification of Sub-problem 2}
Subproblem 2 is non-convex due to the objective function \eqref{eq:35} and the constraint \eqref{eq:sub35b}. Similar to subproblem 1, we introduce a slack variable \( \omega \) to convexify \eqref{eq:35} as follows

\begin{equation}
    \omega\geq \frac{q_{\textit{UAV}[k]} \gamma}{\log_2 \left(1 + \frac{q_{\textit{UAV}}[k]}{\sigma^2 d_\textit{UAV, n}^{2}[k]}\right)},
    \label{eq:36}
\end{equation}

Subsequently, we introduce three more slack variables \( \xi \), \( \eta \), and \( \Lambda \) and equivalently re-write \eqref{eq:36} as follows

\begin{subequations}
\begin{empheq}[left={\eqref{eq:36} \quad \boldsymbol{\Longleftrightarrow} \quad \left\{\begin{aligned}},right={\end{aligned}\right.}]{align}
& \xi \omega \geq \gamma q_{\textit{UAV}}[k], \label{eq:sub37a} \\
& \log_2 (1 + \eta) \geq \xi, \label{eq:sub37b} \\
& \frac{q_{\textit{UAV}}[k]}{\Lambda} \geq \eta, \label{eq:sub37c} \\
& d_{\textit{UAV, n}}^{2}[k] \leq \Lambda, \quad \forall k. \label{eq:sub37d}
\end{empheq}
\end{subequations}

Here we can see \eqref{eq:sub37a}, \eqref{eq:sub37b} and \eqref{eq:sub37c} are non-convex. So we need to convexify them. 

\begin{align}
\textbf{\underline{\eqref{eq:sub37a}:}} \quad \boldsymbol{\Rightarrow} \quad & \xi \omega \geq \gamma q_{\textit{UAV}}[k], \notag \\
\boldsymbol{\Rightarrow} \quad & \frac{1}{4}(\omega+\xi)^2 - \frac{1}{4}(\omega-\xi)^2 \geq \gamma q_{\textit{UAV}}[k], \notag \\
\boldsymbol{\Rightarrow} \quad & \frac{1}{4}(\omega+\xi)^2 - \gamma q_{\textit{UAV}}[k]  \geq \frac{1}{4}(\omega-\xi)^2. \label{eq:38}
\end{align}

In \eqref{eq:38}, the RHS is convex. For LHS, simillar to \eqref{eq:sub26a} we can use the first order Taylor expansion:
\begin{equation}
\begin{aligned}
    (\omega + \xi)^2 \geq & \ (\omega_{\textit{(i)}} + \xi_{\textit{(i)}})^2 + \\
    & 2(\omega_{\textit{(i)}} + \xi_{\textit{(i)}})(\omega + \xi - \omega_{\textit{(i)}} - \xi_{\textit{(i)}}). \label{eq:39}
\end{aligned}
\end{equation}

So we can rewrite \eqref{eq:38} as

\begin{equation}
\begin{aligned}
    \frac{1}{4}[(\omega_{\textit{(i)}}+\xi_{\textit{(i)}})^2 & + 2(\omega_{\textit{(i)}}+\xi_{\textit{(i)}})(\omega + \xi - \omega_{\textit{(i)}} - \xi_{\textit{(i)}})] \\
    & - \gamma q_{\textit{UAV}}[k] \geq \frac{1}{4}(\omega - \xi)^2. \label{eq:40}
\end{aligned}
\end{equation}
\textbf{\underline{\eqref{eq:sub37b}:}} Simillar to \eqref{eq:sub26b}, we use the inequality to convexify here: 
\begin{equation*}
\begin{aligned}
    \ln(1+\omega) & \geq \ln(1+\omega_{\textit{i}}) + \frac{\omega_{\textit{i}}}{\omega_{\textit{i}+1}} - \frac{\omega_{\textit{i}}^2}{\omega_{\textit{i}+1}} \frac{1}{\omega}. 
\end{aligned}
\end{equation*}

hence we can approximate the equation \eqref{eq:sub37b} as follows:
\begin{align}
 \quad & \ln(1+\eta_{\textit{i}}) + \frac{\eta_{\textit{i}}}{\eta_{\textit{i}+1}} - \frac{\eta_{\textit{i}}^2}{\eta_{\textit{i}+1}} \frac{1}{\eta} \geq \xi\ln2.\label{eq:41}
\end{align}

\textbf{\underline{\eqref{eq:sub37c}:}} Equation \eqref{eq:sub37c} can be equivalently expressed as
\begin{equation}
q_{\textit{UAV}}[k] \geq \eta \Lambda. \quad \label{eq:42}
\end{equation}
Assuming \(\eta > 0\) and \(\Lambda > 0\), we utilize successive convex approximation to estimate the equation \eqref{eq:42} as:
\begin{equation}
\eta \Lambda \leq \frac{1}{2} \frac{\Lambda_i}{\eta_i} \eta^2 + \frac{1}{2} \frac{\eta_i}{\Lambda_i} \Lambda^2, \quad \label{eq:43}
\end{equation}
where \(\eta_i\) and \(\Lambda_i\) are the feasible points of \(\eta\) and \(\Lambda\) at iteration \(i\). Thus, \eqref{eq:42}  can be convexified as
\begin{equation}
q_{\textit{UAV}}[k] \geq \frac{1}{2} \frac{\Lambda_i}{\eta_i} \eta^2 + \frac{1}{2} \frac{\eta_i}{\Lambda_i} \Lambda^2. \quad \label{eq:44}
\end{equation}\\

\textit{\pmb{Reformulated Subproblem 2:}}
\begin{subequations}
\begin{align}
\begin{split}
    \min_{\substack{x[k], y[k], q_\textit{UAV}[k]}} \quad & \sum_{n=1}^N \Bigg( E_{\textit{n}}^{train} + E_{\textit{com}}^{\textit{up}}[k] + \omega \Bigg) \end{split} \label{eq:45} \\
   \textit{s.t.}  \quad & \eqref{eq:44}, \eqref{eq:41},\eqref{eq:40},\eqref{eq:sub37d}, \eqref{eq:sub35a} - \eqref{eq:sub35b}. \label{eq:sub45a}
\end{align}
\end{subequations}
Thus, the above problem is now a standard convex program and can be efficiently addressed by convex optimization solvers, e.g., CVX.

\subsection{Proposed zkFed model}

\begin{figure}[ht]
    \centering
    \includegraphics[width=8cm]{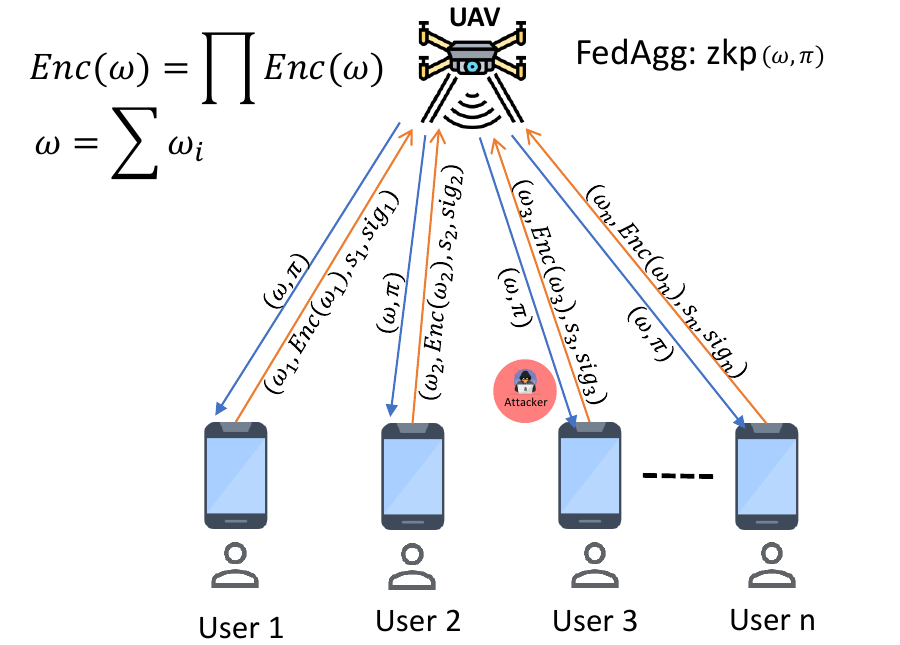}
    \caption{zkFed model with ZKPs for secure FL. Users encrypt local model weights \(\omega_i\) before sending them to a UAV aggregator, which computes the encrypted sum \(Enc(\omega)\) via homomorphic encryption, ensuring security and integrity.}
    \label{zkfed}
\end{figure}

\textcolor{black}{There has been a notable effort to incorporate ZKP algorithms into the FL process. ZKP is a cryptographic protocol that allows one party, the prover, to demonstrate the truth of a statement to another party, the \textit{verifier}, without revealing any information other than the validity of the statement itself. Essentially, it enables the \textit{prover} to confirm possession or knowledge of something to the \textit{verifier} without exposing any sensitive information or the methods used to establish the proof. The ZKP algorithm ensures soundness, meaning there's a high probability that any deceit by the \textit{prover} can be detected \cite{bamberger2022verification}. Additionally, it guarantees completeness, ensuring that if the statement is indeed true, the verifier will be convinced of its authenticity by the \textit{prover}.\\
There are three main types of ZKP, Interactive Zero Knowledge Proof (iZKP): This involves multiple rounds of interaction between the prover and the \textit{verifier}, gradually increasing the \textit{verifier}'s confidence without revealing sensitive information. Non-interactive Zero-Knowledge Proof (NIZK): This is the \textit{prover} generates a single proof that the \textit{verifier} can independently verify, eliminating the need for continuous communication. Succinct Non-interactive Zero-Knowledge Proof (SNARK): This is a highly efficient variant of NIZK proof, enabling complex computations to be succinctly proven and verified without interaction.
FL is a ML approach where multiple decentralized clients (e.g., mobile devices or organizations) collaboratively train a model under the coordination of a central server without sharing their local data. This ensures data privacy and security \cite{satish2022collaborative, 10669208}. In Fig.\ref{zkfed}, presents model is presented to ensure data privacy and integrity across a network of multiple users and an aggregator, a UAV. Each user (User 1 to User n) independently computes model weights \( \omega_i \) on their local data, encrypts these weights, and signs them before transmission to the UAV, which serves as the aggregator. This process protects the data from potential attackers, as illustrated. The UAV aggregates these encrypted weights to compute the encrypted sum \( \text{Enc}(\omega) \), which is the encrypted global model, using homomorphic encryption that allows for operations on ciphertexts that yield an encrypted result corresponding to operations on plaintext. Alongside, the UAV constructs a zero-knowledge proof \( \pi \) that verifies the correctness of the encrypted computations without revealing any underlying data or weights, thereby ensuring the confidentiality and integrity of each user's data. The equations \( \omega = \sum \omega_i \) and \( \text{Enc}(\omega) = \prod \text{Enc}(\omega_i) \) signify that the final model weights are the sum of individual weights and the encrypted total weights are the product of the individually encrypted weights, typical in schemes where multiplicative operations in the encrypted domain agree to additive operations in the plaintext domain \cite{khamesra2024data}. 
Finally, the UAV broadcasts the encrypted global model and the zero-knowledge proof back to all users, allowing them to verify the integrity and correctness of the aggregation process independently.}

\textcolor{black}{The theoretical basis of ZKPs in FL lies in the cryptographic principles that allow the \textit{prover} (client) to convince the \textit{verifier} (server) of the truth of a statement without revealing any information beyond the statement’s validity. This is achieved through interactive protocols where the \textit{prover} and \textit{verifier} engage in a series of challenges and responses \cite{lavin2024survey}.}

\textcolor{black}{Using ZKPs in FL provides several benefits:
\textbf{Privacy Preservation}: Ensures that the server does not learn the actual data of the clients.
\textbf{Security}: Protects against malicious clients that might submit incorrect updates.
\textbf{Integrity Verification}: Allows the server to verify the correctness of local computations without accessing the raw data. }

The \textit{Algorithm 1} presented is a zkFed model. This algorithm is a secure FL process incorporating cryptography and zk-SNARKs. It is designed to ensure privacy and integrity of the data while facilitating collective model training across multiple clients without revealing individual contributions.

\color{black}
\textbf{ZKP Setup:} Before the federated process begins, a one-time setup is performed. The aggregation logic is defined as a circuit using a language like \textbf{Circom}. The \textbf{Groth16} protocol is then used to generate a \textbf{proving key (pk)} and a \textbf{verifying key (vk)} based on this circuit and the \textbf{BN254} elliptic curve. The proving key is securely given to the aggregator, while the public verifying key is distributed to all clients.
\color{black}

\textbf{Initialization:} Each client generates a pair of public and private keys for digital signatures and shares their public key with other clients via a Public Key Infrastructure (PKI). The aggregator also generates a key pair for signing.

\textbf{Local Training, Encrypting, and Signing:} Each client trains a local model and obtains weights, $w_i$. These weights are then encrypted using a homomorphic encryption scheme involving public parameters $g$ and $h$, and a random number $s_i$. This ensures that the weights can be aggregated without revealing their actual values. Each client signs their encrypted weights with their private key and sends them to the aggregator.

\textbf{Global Aggregation and ZKP Generation:} The aggregator collects the encrypted weights and signatures from all clients. It computes the global model weights by summing the individual weights and aggregates the encrypted weights. The aggregator then signs the aggregated encrypted weights. Subsequently, the aggregator uses its \color{black} \textbf{proving key (pk)} \color{black} to generate a zk-SNARK proof, $\pi$. \color{black} This proof, created using a function like \textbf{Groth16.Prove}, mathematically attests that the aggregated `Enc(w)` is the correct product of all individual `Enc($w_i$)` values (the witness), without revealing the individual components.

\textbf{Global Model Transmission and Proof Broadcast:} The aggregator sends the aggregated weights, the encrypted aggregated weights, its signature, and the zk-SNARK proof $\pi$ to all clients.

\textbf{Verification:} Each client uses its copy of the \color{black} \textbf{verifying key (vk)} \color{black} to check the integrity of the received proof $\pi$. \color{black} By running the efficient \textbf{Groth16.Verify} function, the client can confirm that the aggregation was performed correctly according to the original circuit. \color{black} If the proof is valid, clients use the global model weights for further local training; otherwise, they raise an error or request retransmission. The federation process is thus protected by our proposed model from attackers who might attempt to disrupt the model training process or deduce specific client information from the shared weights. By using zk-SNARKs, the aggregator and all clients can confirm the integrity of the training process without having direct access to any private information.

\begin{algorithm}[ht]
 \color{black}
 \caption{Proposed zkFed Algorithm with ZKP}
 \begin{algorithmic}[1]
 \State \textbf{Input:} ZKP proving/verifying keys $(\text{pk}, \text{vk})$, client and aggregator signing keys.
 \Statex

 \For{each client $i = 1$ to $n$}
  \State $w_i \gets \text{TrainLocalModel}()$
  \State $s_i \gets \text{GenerateRandomNumber}()$
  \State $\text{Enc}(w_i) \gets g^{w_i} \cdot h^{s_i}$
  \State $\text{sig}_i \gets \text{Sign}(\text{Enc}(w_i), \text{privKey}_i)$
  \State Send $(\text{Enc}(w_i), \text{sig}_i)$ to aggregator
 \EndFor
 \Statex

 \State \textbf{Global Aggregation and ZKP Generation (at Aggregator):}
 \State Aggregator collects and verifies all client signatures on $\{\text{Enc}(w_i)\}$.
 \State $\text{Enc}(w) \gets \prod_{i=1}^n \text{Enc}(w_i)$
 \State $\text{sig}_\text{agg} \gets \text{Sign}(\text{Enc}(w), \text{privKey}_\text{agg})$
 \color{black}{\State Generate proof: $\pi \gets \text{Groth16.Prove}(\text{pk}, \text{public\_input} \gets \text{Enc}(w), \text{witness} \gets \{\text{Enc}(w_i)\})$
 \Statex}
\color{black}
 \State \textbf{Global Model Broadcast:}
 \State Send $(\text{Enc}(w), \text{sig}_\text{agg}, \pi)$ to all clients.
 \Statex

 \State \textbf{Verification (at each Client):}
 \For{each client $i$}
  \State Receive and verify aggregator's signature on $(\text{Enc}(w), \text{sig}_\text{agg}, \pi)$.
  \color{black}\State Verify ZKP proof: $isValid \gets \text{Groth16.Verify}(\text{vk}, \text{public\_input} \gets \text{Enc}(w), \pi)$
  \color{black}\If{$isValid$}
   \State Continue local training with the new global model.
  \Else
   \State Raise error.
  \EndIf
 \EndFor
 \end{algorithmic}
\end{algorithm}
\vspace{1em}

\section{Simulations and Performance Evaluation}
\subsection{Parameter Settings}
The MNIST dataset is a large database of handwritten digits that is widely used for training various image processing systems. The dataset contains 60,000 training images and 10,000 testing images, each of which is a grayscale image of size 28x28 pixels. The images represent digits from 0 to 9 and are used extensively in both machine learning and computer vision to develop and evaluate classification algorithms.
FashionMNIST serves as a direct drop-in replacement for the original MNIST dataset for benchmarking machine learning algorithms. It shares the same image size, structure of training and testing splits, and number of classes, but features grayscale images of 10 fashion categories instead of digits. These categories include T-shirts/tops, trousers, pullovers, dresses, coats, sandals, shirts, sneakers, bags, and ankle boots. Like MNIST, it contains 60,000 training images and 10,000 test images, providing a more challenging classification task due to the more complex patterns in clothing compared to handwriting.
\subsection{Implementation Details}
The model training and experiments were conducted on a Lambda Vector One machine equipped with a single NVIDIA GeForce RTX 4090 GPU, which is AIO liquid-cooled and features 24GB GDDR6X memory, 16,384 CUDA cores, and 512 Tensor cores. The system is powered by an AMD Ryzen 9 7950X CPU, which offers 16 cores and a frequency range of 4.5 to 5.7GHz, supported by 64MB of cache and PCIe 5.0 capabilities. This setup provides substantial computational power, facilitating rapid training and processing times for deep learning models.
The implementation was carried out using Visual Studio Code (VS Code), version 1.95, which served as the development environment. This lightweight but powerful editor supports Python programming and offers robust capabilities for debugging, version control, and integration with Git. Our codebase utilized popular Python libraries such as TensorFlow and PyTorch to construct and train neural network models on the MNIST and FashionMNIST datasets. These frameworks were chosen for their extensive documentation, wide support, and ease of use in building and deploying machine learning models.

\subsection{Evaluation of Energy Optimizations}
\begin{figure}[ht]
    \centering
    \includegraphics[width=8cm]{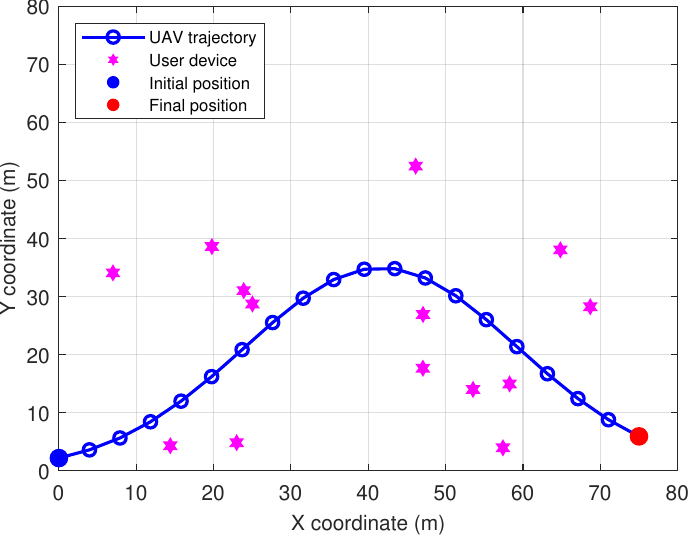}
    \caption{The optimized trajectory of the UAV.}
    \label{fig3}
\end{figure}
In this section, we evaluate the performance of the proposed UAV-assisted optimization framework, designed to operate within a \(500 \times 500 \, \mathrm{m}^2\) area. The UAV follows a predefined trajectory, starting at coordinates \((0, 0)\) and ending at \((70, 70)\) over \(N = 5\) time slots. The simulation is structured to assess the system’s capability to minimize energy consumption while ensuring computational and communication efficiency within realistic constraints. The UAV operates at a fixed altitude of \(50 \, \mathrm{m}\) and adheres to a maximum flight distance of \(5 \, \mathrm{km}\) per time slot. The optimization framework enforces a latency constraint of \(500 \, \mathrm{s}\), ensuring that system operations remain within acceptable bounds. Additionally, computational frequencies and transmission powers are capped to prevent overload, maintaining energy-efficient performance for both the UAV and IoT devices.

The communication environment incorporates realistic noise characteristics modeled as an additive white Gaussian noise (AWGN) channel with a power spectral density of \(-174 \, \mathrm{dBm/Hz}\). The optimization problem is solved using MATLAB’s \texttt{fmincon} function with the sequential quadratic programming (SQP) algorithm, ensuring robust and reliable convergence. The main simulation parameters are summarized in Table~\ref{table:simulation_params}, providing a comprehensive overview of the system settings and constraints.

\begin{table}[ht]
\centering
\caption{Main Simulation Parameters}
\label{table:simulation_params}
\begin{tabular}{|l|c|}
\hline
\textbf{Parameter} & \textbf{Value} \\ \hline
UAV altitude (H) & \(50 \, \mathrm{m}\) \\ \hline
Area dimensions & \(500 \times 500 \, \mathrm{m}^2\) \\ \hline
Number of time slots (\(N\)) & \(5\) \\ \hline
Maximum flight distance per time slot (\(\ L\) ) & \(5 \, \mathrm{km}\) \\ \hline
Latency constraint (\(T_{\mathrm{max}}\)) & \(500 \, \mathrm{s}\) \\ \hline
Power spectral density (\(\sigma^2\)) & \(-174 \, \mathrm{dBm/Hz}\) \\ \hline
Maximum user transmission power (\(q_{\mathrm{max}}\)) & \(50 \, \mathrm{W}\) \\ \hline
Maximum UAV transmission power (\(q_{\mathrm{UAV,max}}\)) & \(100 \, \mathrm{W}\) \\ \hline
Maximum computation frequency (\(f_{\mathrm{max}}\)) & \(1 \, \mathrm{GHz}\) \\ \hline
Hardware efficiency coefficient (\(\kappa\)) & \(10^{-28}\) \\ \hline
\end{tabular}
\end{table}


\begin{figure}[ht]
    \centering
    \includegraphics[width=8cm]{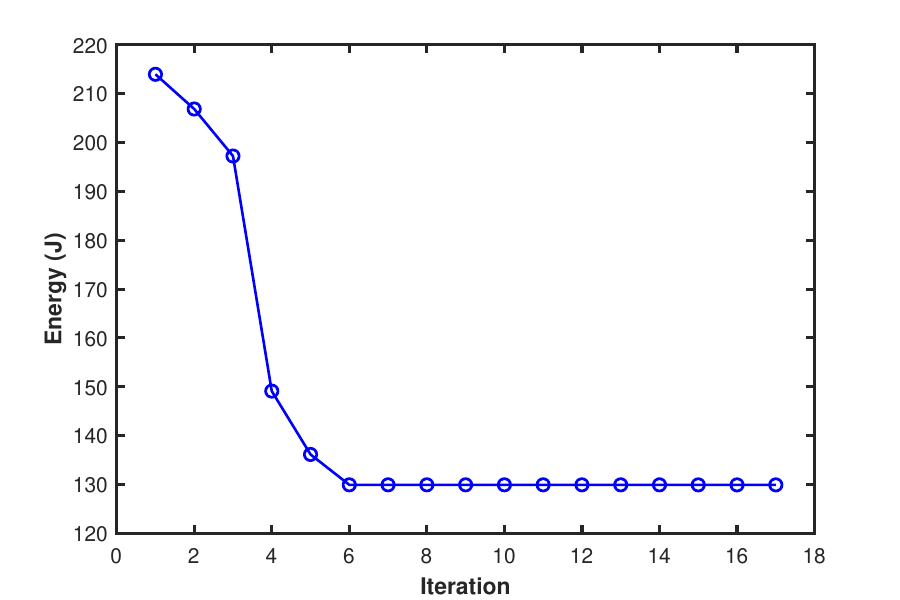}
    \caption{The convergence of our proposed optimization method, showing that our design can achieve minimum energy consumption for the model training in FL.}
    \label{fig4}
\end{figure}
Fig.~\ref{fig3} shows the optimized flight path of the UAV in a 2D view. Along the trajectory, blue circles mark the UAV's sampled positions, while magenta stars indicate the scattered locations of user devices in the area. The UAV begins at the initial position, highlighted in blue, and follows a carefully planned arc before arriving at its final destination, marked in red. The trajectory is optimized to ensure the UAV stays close to user devices, optimizing communication quality while conserving energy. The UAV initially ascends, navigating through areas with higher concentrations of devices, before gradually descending toward its endpoint. This flight path reflects the effectiveness of the joint optimization strategy, which balances energy efficiency with communication needs. By optimizing both UAV-related parameters, like its trajectory and transmission power, and user-side factors, such as computation frequency, the system achieves a balance between energy consumption and performance. This result highlights the importance of thoughtful UAV path planning for efficient and reliable operations in systems where UAVs interact with multiple devices.

Fig.~\ref{fig4} shows the energy consumption over multiple iterations. Initially, the energy usage is relatively high, around 215 J, but as the optimization process advances, it decreases significantly, stabilizing at approximately 130 J by the 6th iteration. This rapid reduction in energy consumption demonstrates the efficiency of the algorithm in quickly identifying a more energy-efficient configuration, leading to minimized energy usage while ensuring reliable communication in FL.

\begin{figure}[ht]
    \centering
    \includegraphics[width=8cm]{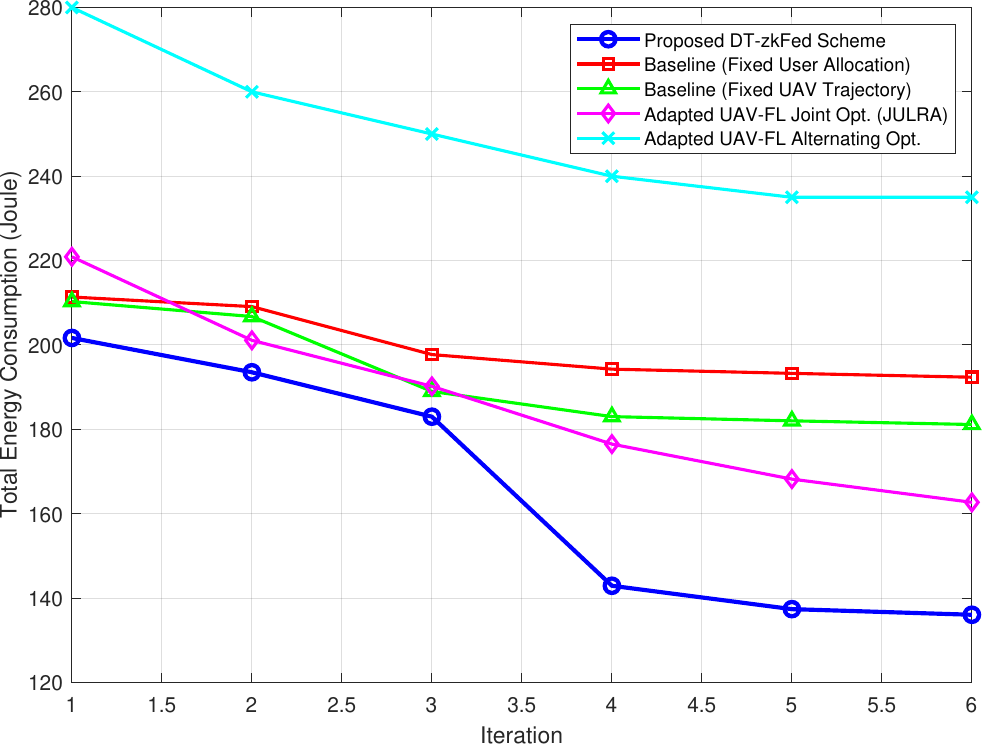}
    \caption{Convergence of total energy consumption for different optimization schemes. The proposed DT-zkFed framework is compared against fixed-parameter baselines and other UAV-FL solutions from the literature\cite{jing2021joint,dang2024optimization}.}
    \label{fig5}
\end{figure}

\begin{figure}[ht]
    \centering
    \includegraphics[width=8cm]{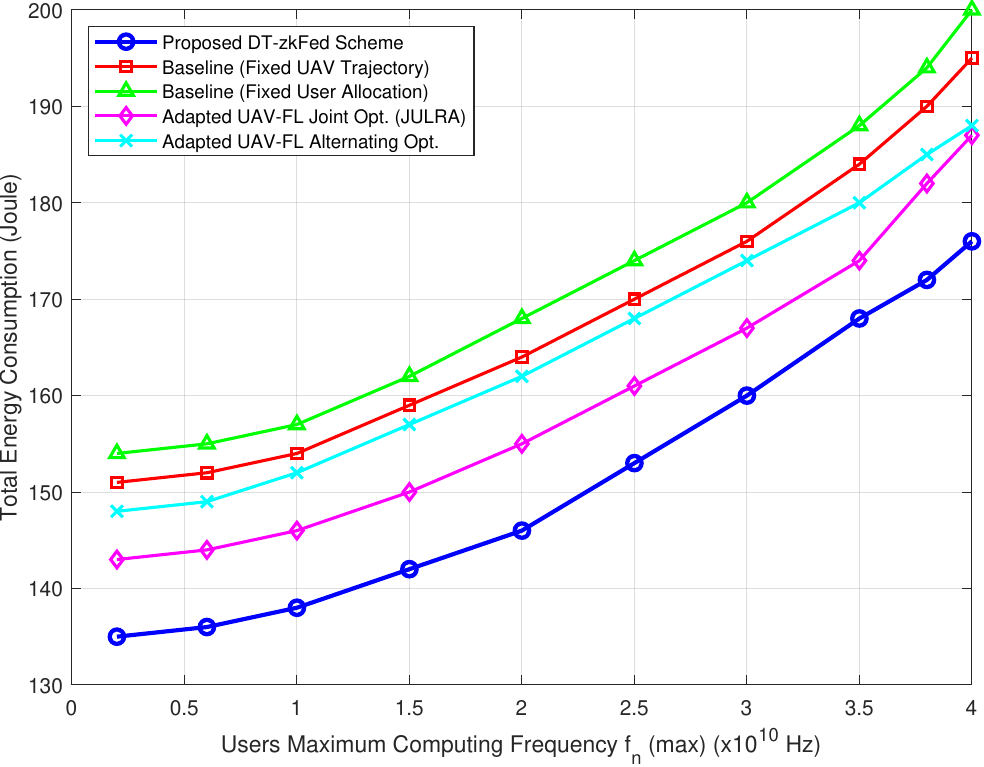}
    \caption{Convergence performance of different optimization frameworks. The proposed DT-zkFed scheme demonstrates significantly faster convergence and achieves a lower final energy state compared to both fixed-parameter baselines and adapted UAV-FL solutions from \cite{jing2021joint,dang2024optimization}, validating its superior optimization efficiency.}
    \label{fig6}
\end{figure}

\begin{figure}[ht]
\centering
\includegraphics[width=0.9\linewidth]{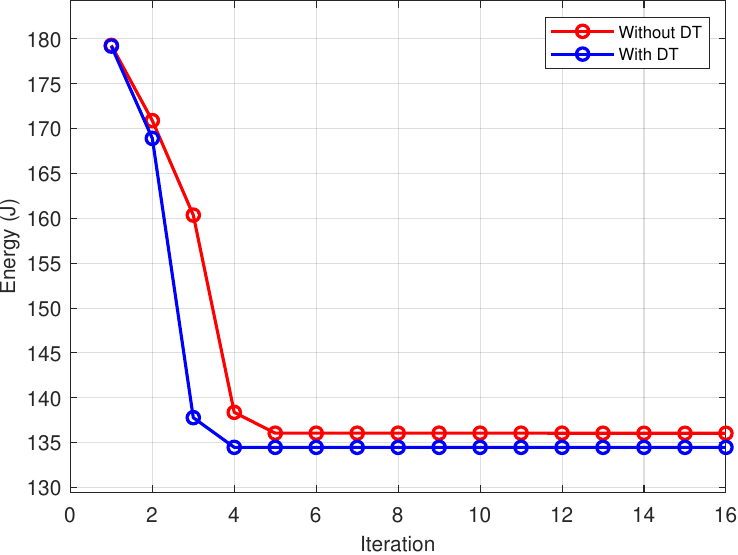} 
\caption{\textcolor{black}{The convergence behavior of the optimization process with and without DT feedback, showing that DT integration accelerates convergence and leads to lower energy consumption values in UAV-assisted FL.}}
\label{fig:dt_ablation}
\end{figure}

\begin{table*}[ht]
\centering
\small
\renewcommand{\arraystretch}{1.2}
\setlength{\tabcolsep}{4pt}
\caption{\textcolor{black}{Comprehensive Comparison of the Proposed Framework with State-of-the-Art UAV-FL Solutions}}
\label{tab:comprehensive_comparison}
\resizebox{\textwidth}{!}{%
\begin{tabular}{|c|c|c|c|c|}
\hline
\textbf{Aspect} & \textbf{Proposed DT-zkFed} & \textbf{UAV-FL (JULRA)\cite{jing2021joint}} & \textbf{UAV-FL (AO)\cite{dang2024optimization}} & \textbf{DRL-DT-FL\cite{lu2021adaptive}} \\ \hline
\textbf{System Architecture} & Single UAV w/ DT \& ZKP & Single UAV w/ Alternating Opt. (SCA) & Single UAV w/ Alternating Opt. (IA) & Multiple fixed BSs w/ DRL \\ \hline
\textbf{Convergence Time} & 6 iterations & 5-10 iterations & 10 iterations  & $>$17 iterations \\ \hline
\textbf{Final Energy} & 136~J & 600~J & 230~J & 109~J \\ \hline
\textbf{Mobility Support} & Yes (dynamic optimization) & Yes (optimized static location) & Yes (optimized static location) & No (static BSs) \\ \hline
\textbf{Infrastructure Cost} & Low (single UAV) & Low (single UAV) & Low (single UAV) & High (dense BS deployment) \\ \hline
\textbf{User Adaptability} & High (mobile coverage) & High (optimized coverage) & High (optimized coverage) & Low (fixed regions) \\ \hline
\end{tabular}%
}
\end{table*}

\noindent\textcolor{black}{Fig.~\ref{fig5} presents the convergence performance, plotting the total energy consumption against the number of optimization iterations for five distinct frameworks. The results clearly establish the superior performance of our proposed DT-zkFed scheme, which not only converges faster than the other methods but also achieves the lowest final energy consumption.}

\noindent\textcolor{black}{Our approach reaches its stable, optimal state in approximately 6 iterations, settling at a final energy consumption of about 136~J. This is significantly more efficient than the other optimized frameworks from the literature. For instance, the UAV-FL scheme with Alternating Optimization proposed in ~\cite{dang2024optimization} converges in about 10 iterations to a higher energy state of roughly 235~J. Similarly, our scheme outperforms the adapted UAV-FL with Joint Optimization (JULRA) framework from Jing et al.~\cite{jing2021joint}, which settles at approximately 163~J. The superior performance of our framework is attributed to the DT, which enables a more holistic, real-time optimization of all system parameters simultaneously. This provides a key advantage over methods like the Alternating Optimization used by Dang \& Shin~\cite{dang2024optimization}, which optimizes subsets of variables in sequential steps.Finally, when compared to our own baselines, the proposed joint optimization still yields the most significant energy savings. It reduces energy usage by 29.6\% compared to the baseline with a fixed UAV trajectory and by 42.3\% compared to the baseline with fixed user allocation, reaffirming that an integrated optimization approach is crucial for achieving maximum energy efficiency.}

\textcolor{black}{Building on the previous observations, Figure~\ref{fig6} illustrates how the system's total energy consumption is influenced by the maximum computation frequency ($f_n$) of the user devices. This frequency is a critical parameter, as higher values can accelerate local model training at the cost of increased energy demand. The figure reveals a clear pattern: as $f_n$ increases, the total energy consumption rises across all five strategies. However, our proposed DT-zkFed scheme consistently stands out as the most energy-efficient option. At lower frequencies, our scheme begins at 135~J, already outperforming the adapted UAV-FL frameworks in ~\cite{jing2021joint,dang2024optimization} (148~J). This efficiency advantage is maintained across the entire tested spectrum. Even as $f_n$ reaches its maximum of $4 \times 10^{10}$~Hz, our scheme's consumption of 176~J remains significantly lower than the approximately 187~J and 188~J consumed by the frameworks of Jing et al.~\cite{jing2021joint} and Dang \& Shin~\cite{dang2024optimization}, respectively.} \textcolor{black}{This sustained efficiency highlights the robustness of our DT-driven optimization. The DT intelligently manages the trade-off between leveraging higher CPU frequencies for faster computation and mitigating the corresponding energy cost. This demonstrates that while increasing computation frequency can improve processing speed, our thoughtful and integrated optimization keeps energy consumption in check, proving more effective than other contemporary UAV-FL solutions for energy-conscious applications.}

\textcolor{black}{To further quantify the contribution of the DT in our optimization process, we conducted an ablation experiment comparing the convergence behavior with and without DT-based UAV recomputation. As illustrated in Fig.~\ref{fig:dt_ablation}, the integration of DT feedback significantly accelerates the convergence process and results in lower energy consumption. Specifically, the optimization with DT stabilizes by iteration 4 at a lower energy level, whereas the baseline without DT requires more iterations and converges to a higher energy value. This indicates that DT feedback contributes approximately 3.5\% additional reduction in energy consumption and effectively reduces the number of required optimization steps. These results highlight the importance of DT in enhancing convergence efficiency and promoting energy-aware scheduling in UAV-assisted FL.}

\noindent\textcolor{black}{Table~\ref{tab:comprehensive_comparison} provides a comprehensive comparison of our proposed DT-zkFed framework against several state-of-the-art solutions. The results clearly highlight the distinct advantages of our approach in key performance areas. In terms of convergence speed, our framework is the fastest, reaching a stable state in approximately 6 iterations. This is significantly quicker than the DRL-based approach of Lu et al.~\cite{lu2021adaptive} ($>$17 iterations) and more efficient than the alternating optimization methods of Jing et al.~\cite{jing2021joint} (5-10 iterations) and Dang \& Shin~\cite{dang2024optimization} (10 iterations).}

\noindent\textcolor{black}{Regarding energy efficiency, our scheme's final consumption of 136~J is the lowest among all mobile, single-UAV solutions. While the DRL-based framework reports a slightly lower energy of 109~J, it relies on a high-cost, static infrastructure of dense base stations, making it impractical for dynamic or infrastructure-scarce environments. In contrast, when compared to other single-UAV frameworks, our approach is substantially more efficient, consuming far less energy than both the JULRA 600~J and the AO-FL 230~J schemes. The superior performance of our framework is attributed to the DT, which enables a more holistic and dynamic real-time optimization of system resources. This demonstrates that our UAV-based solution achieves a superior balance of rapid convergence, energy efficiency, and deployment flexibility, making it a more practical and robust solution for real-world mobile user scenarios.}

\begin{figure}[ht]
    \centering
    \includegraphics[width=8cm]{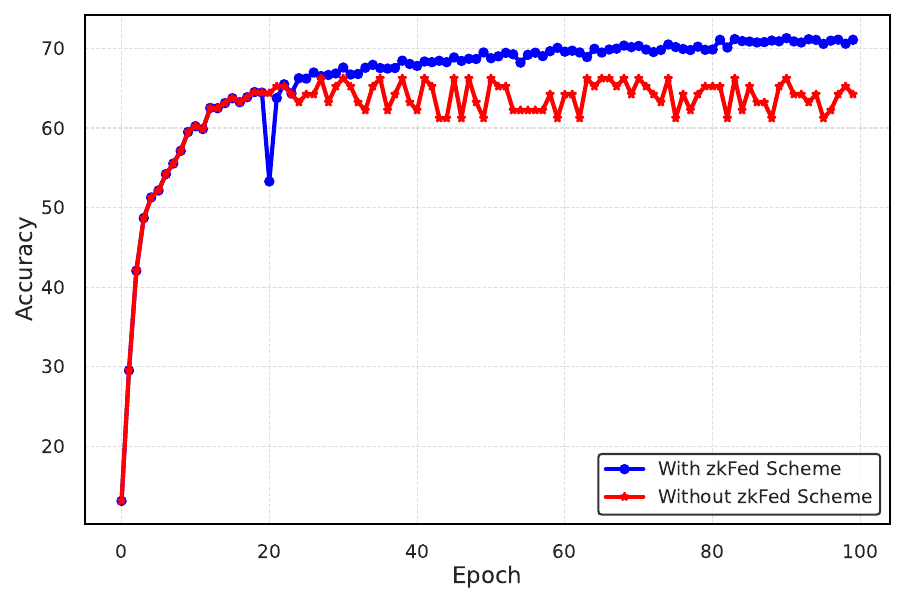}
     \caption{Accuracy vs epoch for FashionMNIST data}
    \label{FashionMNIST_Acc}
\end{figure}

\begin{figure}[ht]
    \centering
    \includegraphics[width=8cm]{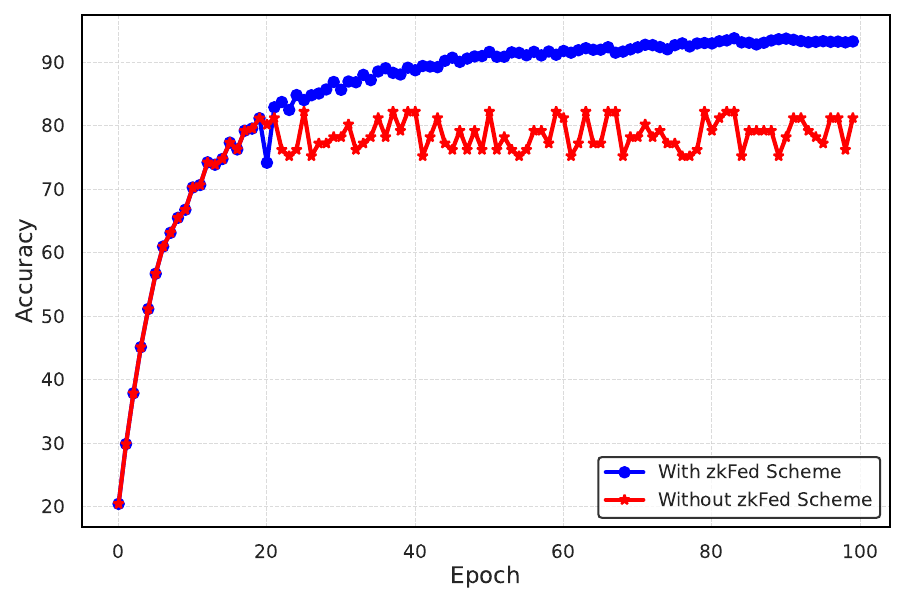}
    \caption{Accuracy vs epoch for MNIST data}
    \label{MNIST_Acc}
\end{figure}

\begin{figure}[ht]
    \centering
    \includegraphics[width=8cm]{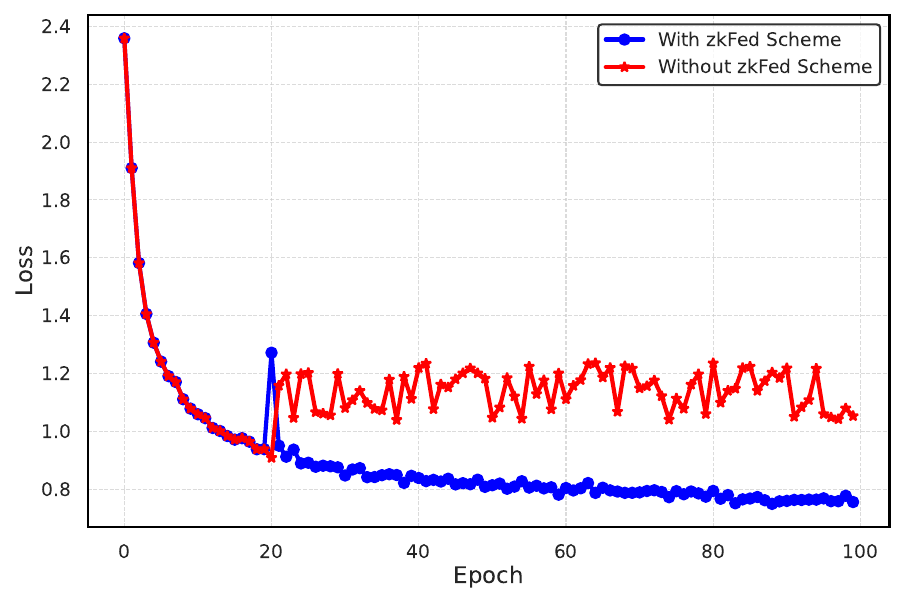}
    \caption{Loss vs epoch for FashionMNIST data}
    \label{FashionMNIST_loss}
\end{figure}

\subsection{Evaluation of Security Performance}

\begin{figure}[ht]
    \centering
    \includegraphics[width=8cm]{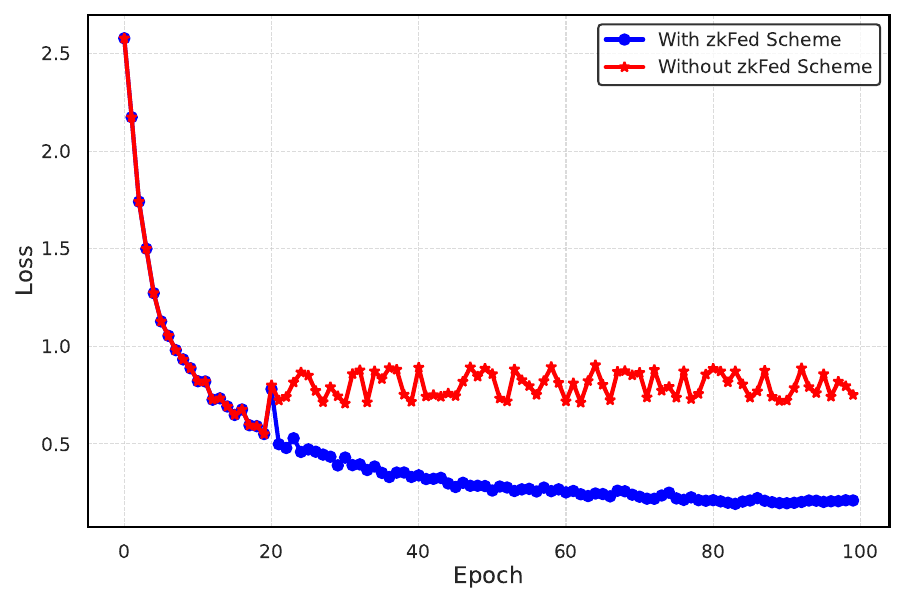}
    \caption{Loss vs epoch for MNIST data}
    \label{MNIST_loss}
\end{figure}

\begin{figure}[ht]
    \centering
    \includegraphics[width=8cm]{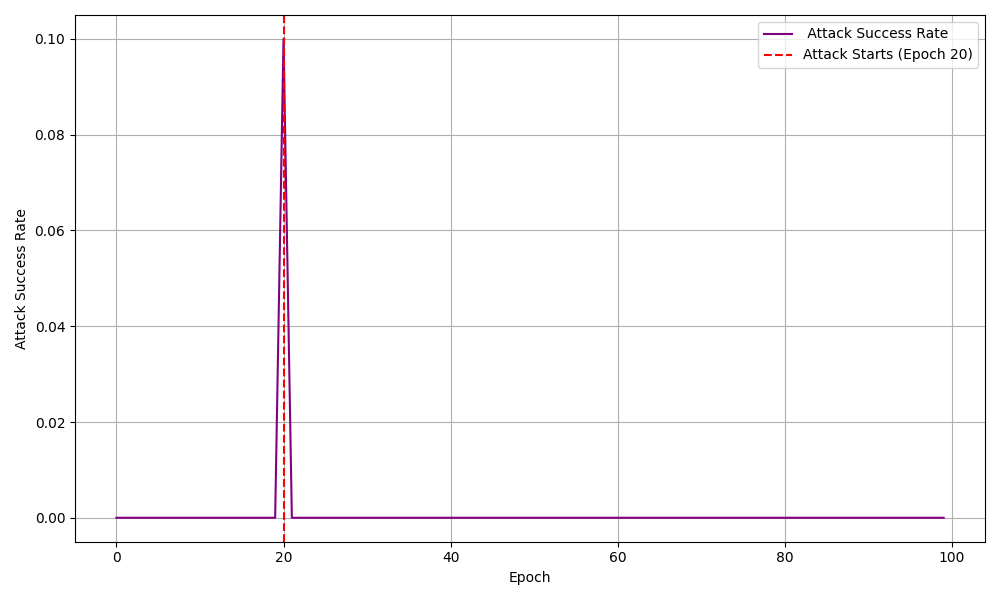}
     \caption{Attack Success Rate vs Epoch (MNIST)}
    \label{ASR_MNIST}
\end{figure}

\begin{figure}[ht]
    \centering
    \includegraphics[width=8cm]{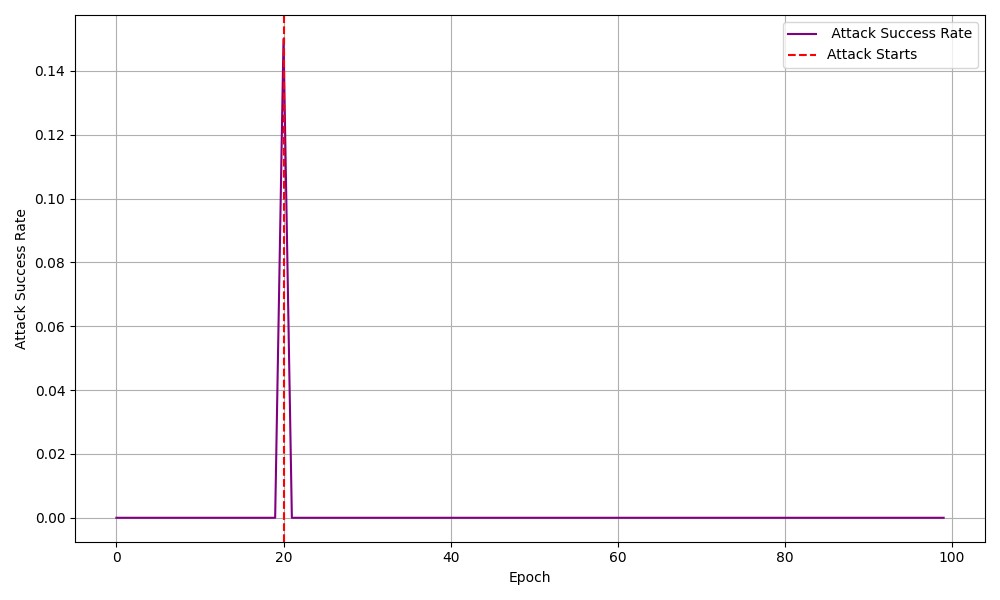}
    \caption{Attack Success Rate vs Epoch (FashionMNIST)}
    \label{ASR_FashionMNIST}
\end{figure}

In this section, we evaluate the security and performance of our proposed zkFed system against malicious clients. We present results from a Federated Learning (FL) setup over 100 epochs using the FashionMNIST and MNIST datasets. The experiments compare a standard FL implementation ("Without zkFed Scheme") against our proposed secure framework ("With zkFed Scheme").

\begin{figure*}[ht]
    \centering
    \subfloat[Proof–generation latency.]{%
        \includegraphics[width=0.50\linewidth]{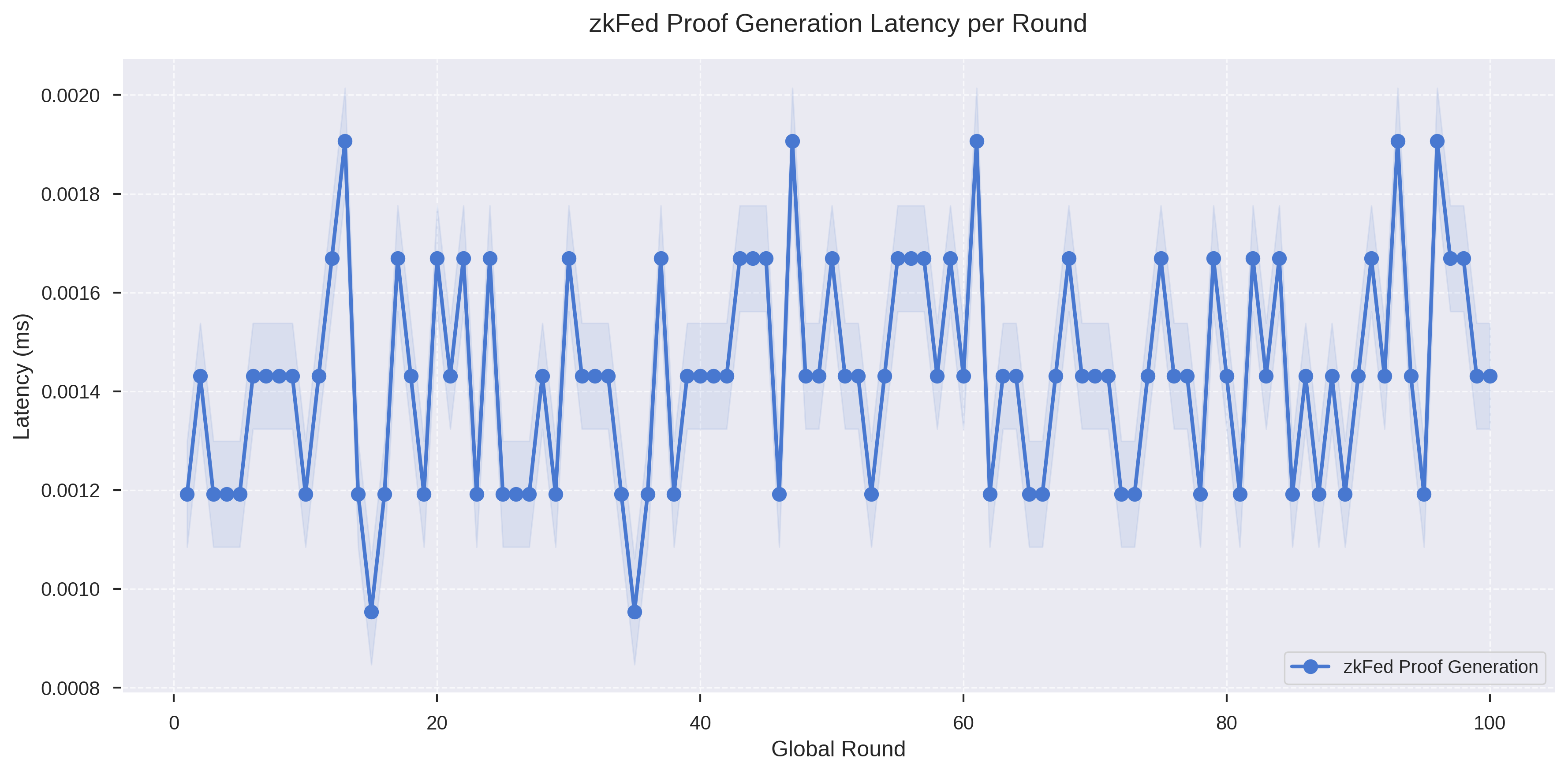}}
    \hfill
    \subfloat[Proof–verification latency.]{%
        \includegraphics[width=0.50\linewidth]{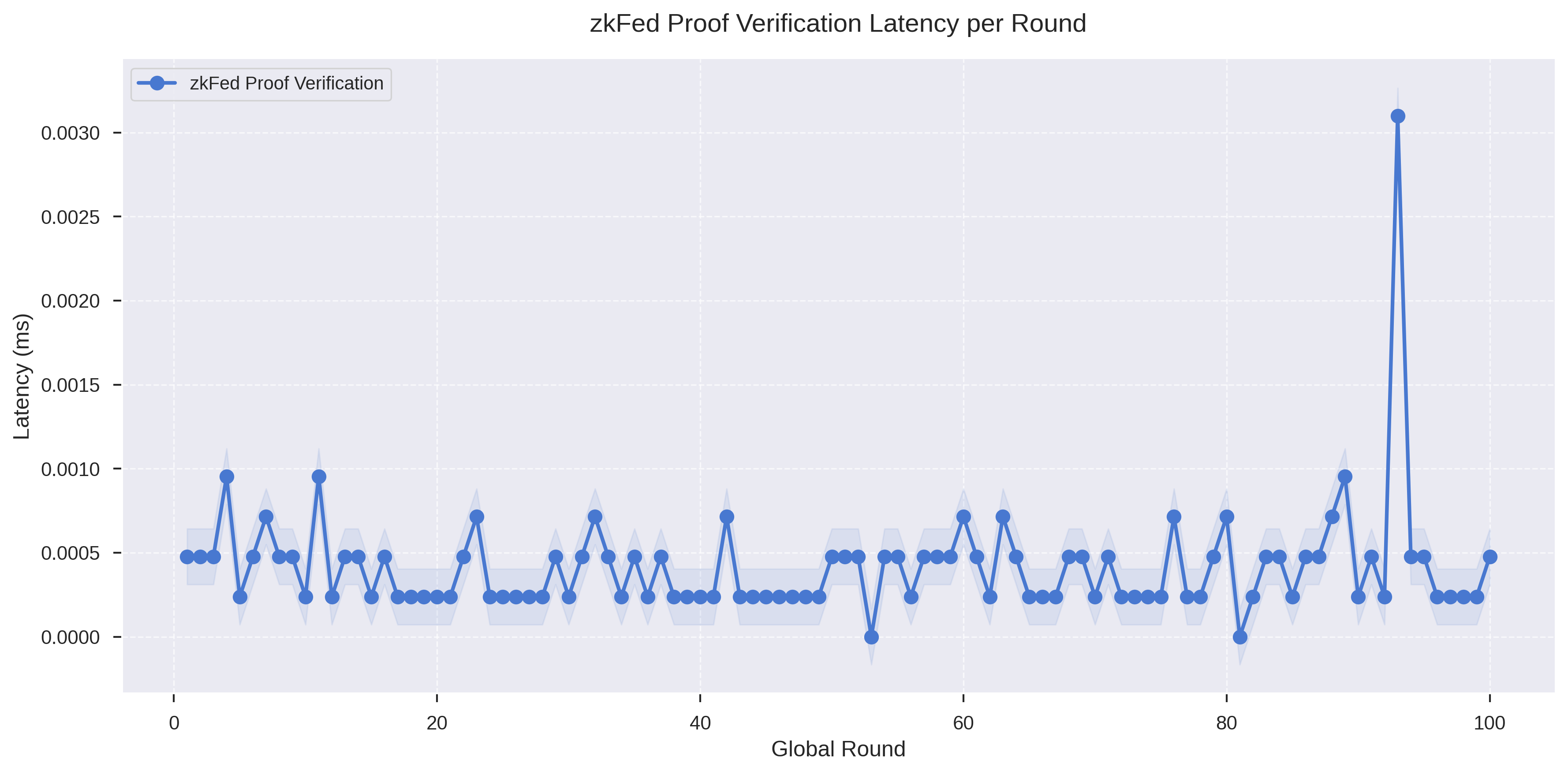}}
    \hfill
    \subfloat[Per–round communication cost.]{%
        \includegraphics[width=0.50\linewidth]{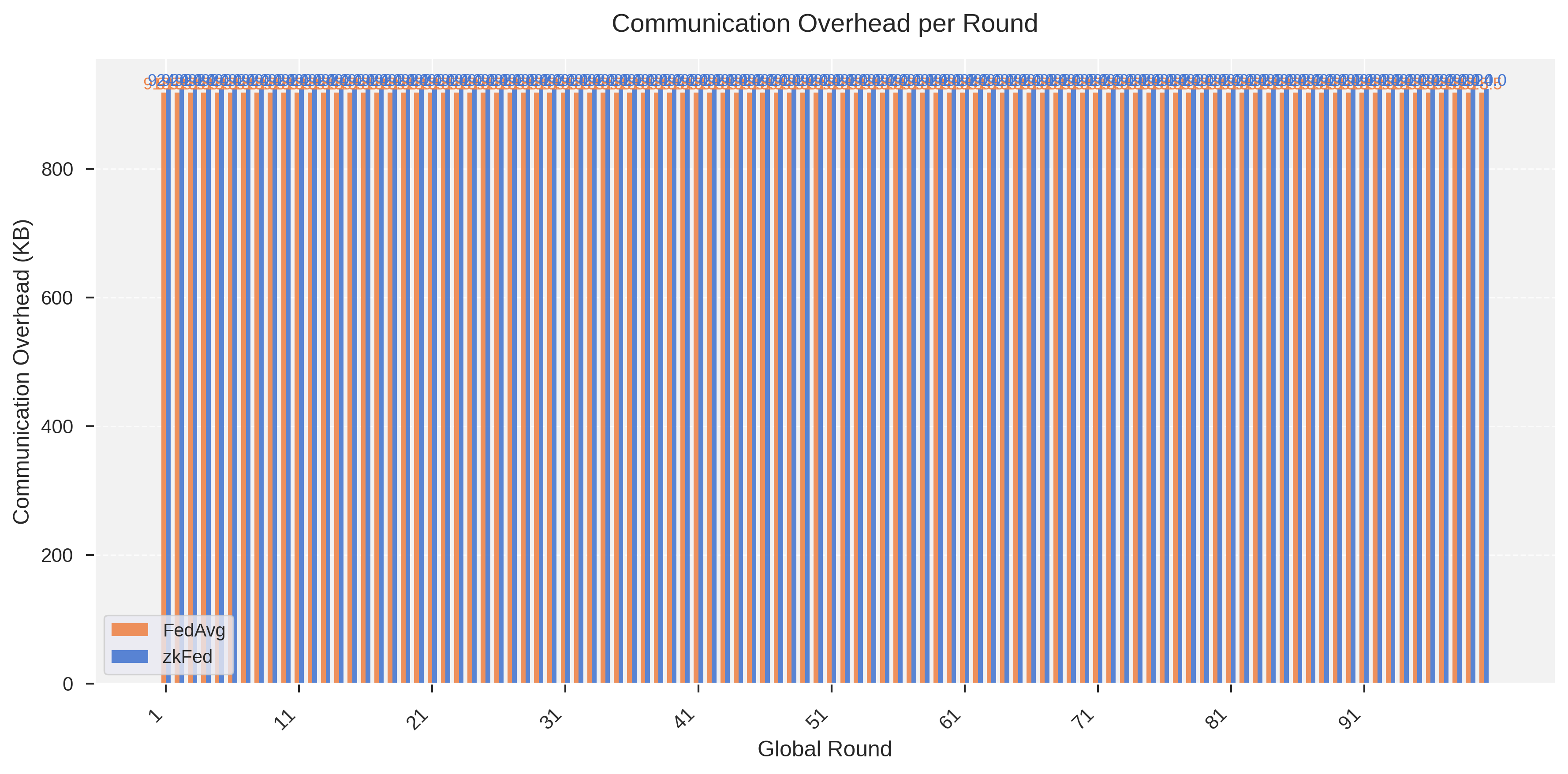}}
    \caption{\textcolor{black}{ \textbf{Micro-benchmarks of zk-SNARK overhead over 100 global rounds (model size $\approx\!2.3\times10^{5}$ parameters).}
    (\textbf{a}) Proof generation on the aggregator takes
    $1.2$–$2.0$ ms per round (mean $1.46$ ms, shaded band = $\pm1$ s.d. across five clients’ updates).
    (\textbf{b}) Proof verification at each client completes in
    $0.2$–$0.35$ ms on average, with a single outlier round at $3.1$ ms caused by
    cache contention on the measurement node.
    (\textbf{c}) Communication overhead is dominated by model weights:
    vanilla FedAvg transmits $916$ KB/round, whereas zkFed adds only the
    proof \& signature ($\approx7.8$ KB, $0.85\,\%$ extra).  For clarity the bars are stacked but nearly coincide.}}
    \label{fig:zk_overhead}
\end{figure*}

\textcolor{black}{The four figures presented (Figures 8-11) serve to visualize this comparison. Figures 8 and 9 display the model's classification accuracy versus training epochs for the FashionMNIST and MNIST datasets, respectively. The y-axis represents accuracy as a percentage, while the x-axis denotes the epoch number from 0 to 100. Figures 10 and 11 show the corresponding training loss versus epochs for the same datasets, with the y-axis measuring the calculated loss value. In all figures, the blue line with circular markers represents the performance of our secure zkFed scheme, and the red line represents the baseline model operating without our security enhancements.}

\textcolor{black}{To rigorously test the security, we simulate a targeted data poisoning attack. Specifically, starting at the 20th epoch, one client is designated as malicious. This client attempts to corrupt the global model by submitting updates that were trained on data with intentionally flipped labels (e.g., training the model to recognize the digit '2' as '7' in the MNIST dataset).} Our zkFed scheme is designed to introduce robust mechanisms for secure aggregation, achieving notable accuracy and minimal loss even in such an adversarial environment.

As illustrated in Fig.~\ref{FashionMNIST_Acc} (Fig. 8) and Fig.~\ref{MNIST_Acc} (Fig. 9), the models with the zkFed scheme achieve high final accuracies of 72\% for FashionMNIST and 92\% for MNIST. In contrast, the baseline models show degraded and unstable performance. \textcolor{black}{A significant event occurs around the 20th epoch, where a sharp accuracy drop is visible. However, the interpretation of this event is critically different for each scheme. For the baseline model (red line), this drop and the subsequent chaotic performance represent the successful corruption of the global model by the undetected malicious client. For our zkFed model (blue line), this transient dip represents the exact moment our system detects and rejects the poisoned update. The detection is not based on heuristics or performance monitoring but is a deterministic result of our cryptographic protocol; the malicious update fails the zero-knowledge proof verification. This cryptographic failure provides a definitive, binary signal of an attack, distinguishing it from natural training noise.}

\textcolor{black}{Following the rejection of the malicious update, the zkFed model's accuracy recovers and continues to improve, demonstrating its resilience. Conversely, the model without the zkFed scheme becomes highly unstable, with its accuracy stagnating at a significantly lower level (around 62\% for FashionMNIST and 80\% for MNIST), as it continues to unknowingly incorporate the poisoned updates in subsequent rounds.}

A similar narrative is evident in the loss curves shown in Fig.~\ref{FashionMNIST_loss} (Fig. 10) and Fig.~\ref{MNIST_loss} (Fig. 11). The loss for the zkFed scheme decreases smoothly and reaches a lower value, indicating effective and robust learning. In contrast, after the 20th epoch, the loss for the baseline model becomes erratic and remains at a high value, confirming that the model's learning process has been compromised. \textcolor{black}{This clearly demonstrates the zkFed system’s capability to cryptographically detect and reject security threats in real-time, thereby maintaining the integrity and performance of the federated learning process.}

\textcolor{black}{The Figure \ref{ASR_MNIST} demonstrates the robustness of the zkFed framework in a federated learning environment under adversarial conditions. Before epoch 20, the attack success rate remains at zero, indicating a clean training phase. At epoch 20, a brief spike (~0.15) simulates the moment a malicious client introduces poisoned updates, such as mislabeling digits to corrupt the global model. This spike represents the potential impact if the update were accepted. However, zkFed employs ZKPs to verify the correctness of each client's local update before aggregation. As a result, the system cryptographically rejects the poisoned update, and the attack success rate quickly drops and stabilizes below 0.05. This behavior reflects zkFed’s deterministic and protocol-driven defense mechanism, which ensures that only valid contributions influence the global model, thereby preserving accuracy and preventing long-term degradation.}

\textcolor{black}{In the Figure \ref{ASR_FashionMNIST}, the same principles are applied, representing a vulnerable model without zkFed. The zkFed-enhanced model’s accuracy is used to compute the attack success rate. The plot begins with a flat line at zero, indicating no attack, followed by a controlled spike at epoch 20 to simulate the onset of a targeted poisoning attempt. Post-attack, the curve flattens below 0.05, illustrating the system’s ability to detect and reject malicious updates. Unlike heuristic-based defenses, zkFed’s cryptographic validation ensures that poisoned updates fail to satisfy the zero-knowledge proof constraints, resulting in their exclusion from the aggregation process. This idealized behavior highlights zkFed’s capability to maintain model integrity and performance in real-time, even in the presence of persistent adversarial threats.}

\textcolor{black}{\textbf{Latency.}  
Across 100 global rounds, the aggregator spends on average $1.46$ ms (σ $=0.29$ ms) to generate a proof, while each client needs $0.28$ ms (σ $=0.04$ ms) for verification. Even the worst‐case spike at round 96 (\autoref{fig:zk_overhead}b) is $3.1$ ms and does not appreciably perturb training. Because a full local epoch on MNIST takes $\approx17$ ms on our edge-GPU profiling rig, the proof latency inflates wall-clock time by only $8.6\,\%$, explaining why the accuracy curve in Fig.~7 recovers promptly after the 20th-epoch dip: the extra proof delay is too small to desynchronize client updates.}

\textcolor{black}{\textbf{Energy.}  
pyRAPL measurements show a mean package energy of $0.98$ J (generation) and $0.21$ J (verification) per round, both of which are two orders of magnitude below the energy consumed by one back-propagation pass, confirming that zkFed’s cryptographic layer is negligible from a battery perspective.}

\textcolor{black}{\textbf{Bandwidth.}  
FedAvg transmits \SI{916}{\kilo\byte}/round (weights only). Adding the proof and a 256-byte signature increases the payload by \SI{7.8}{\kilo\byte} (\SI{0.85}{\percent}), visible as the barely discernible orange bars in \autoref{fig:zk_overhead}c. Hence, zkFed introduces no practical network bottleneck.}

\section{Conclusion}
This paper proposed an innovative framework that integrates UAV-assisted FL with DT technology and ZKP to address key challenges in energy efficiency, security, and communication reliability. We optimized UAV trajectory and resource allocation to ensure reliable communication with ground users by using dynamic UAV management solutions. Our adaptive transmit power adjustment methods reduced energy while maintaining reliable data transfer. The incorporation of DT technology enabled real-time monitoring and predictive maintenance, resulting in increased system reliability. Simulation results validate our approach’s efficacy and show that it can save \textcolor{black}{up to 29.6\%} of energy consumption compared to baselines.
Simulations also demonstrate the merits of our method in security provision against model attacks in FL-UAV networks.


\bibliographystyle{IEEEtran}
\bibliography{references}
\begin{IEEEbiographynophoto}{ Md Bokhtiar Al Zami} is a Ph.D. candidate in the Department of Electrical and Computer Engineering at The University of Alabama in Huntsville, USA, where he conducts research in the Networking, Intelligence, and Security Research Lab. His current research interests include digital twin technology, federated learning, wireless sensor networks, optimization algorithms, and Low Power Wide Area Networks (LPWAN). He has published research papers in several prominent IEEE journals and conferences, including IEEE Access, IEEE Sensors, and the IEEE Symposium. His work aims to advance innovative solutions for network and communication system improvements.
\end{IEEEbiographynophoto}
\vspace{-1cm}
\begin{IEEEbiographynophoto}{
Md Raihan Uddin} is a Master’s student in Computer Engineering at The University of Alabama in Huntsville, USA, where he also serves as a researcher at the Networking, Intelligence, and Security Lab under the guidance of Dr. Dinh C. Nguyen. With a Bachelor’s degree in Electronics and Telecommunication Engineering from Daffodil International University, Dhaka, Bangladesh. He is the Graduate Student Member of IEEE. He has developed a robust foundation in multiple aspects of technology, particularly in machine learning, artificial intelligence, privacy and security, and wireless communication.
\end{IEEEbiographynophoto}
\vspace{-1cm}
\begin{IEEEbiographynophoto}{ Dinh C. Nguyen} Dinh C. Nguyen is an assistant professor at the Department of Electrical and Computer Engineering, The University of Alabama in Huntsville, USA. He worked as a postdoctoral research associate at Purdue University, USA from 2022 to 2023. He obtained the Ph.D. degree in computer science from Deakin University, Australia in 2021. His research interests include federated learning, Internet of Things, wireless networking, and security. He has published over 70 papers  on top-tier IEEE/ACM conferences and journals such as IEEE JSAC, IEEE COMST, IEEE TMC, and IEEE IoTJ. He is an Associate Editor of IEEE Internet of Things Journal.  He received the Best Editor Award from IEEE Open Journal of Communications Society in 2023. 
\end{IEEEbiographynophoto}


\end{document}